\documentclass{article} % For LaTeX2e

\usepackage[table,dvipsnames]{xcolor}

\usepackage{iclr2027_conference,times}

\usepackage{amsmath,amsfonts,bm}

\def\eqref#1{equation~\ref{#1}}
\def\1{\bm{1}}

\DeclareMathAlphabet{\mathsfit}{\encodingdefault}{\sfdefault}{m}{sl}
\SetMathAlphabet{\mathsfit}{bold}{\encodingdefault}{\sfdefault}{bx}{n}

\usepackage{hyperref}
\usepackage{url}
\usepackage{graphicx}
\usepackage{booktabs}
\usepackage{multirow}

\usepackage{amsmath}
\usepackage{url}
\usepackage{tabularx} 
\usepackage{caption}
\usepackage{algorithm}
\usepackage{algorithmic}
\usepackage{float}
\usepackage{enumitem}
\usepackage[skins,breakable]{tcolorbox}
\usepackage{adjustbox}
\usepackage{bbm}
\usepackage{wrapfig}

\title{PORTER: Edge–Cloud Residency for Persistent 3D Scene Graph Memory}
\author{
\makebox[\linewidth][c]{%
{\normalfont Yue Chang\textsuperscript{1} \quad
Yifan Tian\textsuperscript{1} \quad
Jiajing Peng\textsuperscript{2} \quad
Dazhi Huang\textsuperscript{1} \quad
Rufeng Chen\textsuperscript{1} \quad
}%
} \\
\makebox[\linewidth][c]{%
{\normalfont Zhaofan Zhang\textsuperscript{1} \quad
Li Chen\textsuperscript{1} \quad
Sihong Xie\textsuperscript{1,\textdagger}}%
} \\[2pt]
\makebox[\linewidth][c]{%
{\normalfont\small
\textsuperscript{1}The Hong Kong University of Science and Technology
(Guangzhou)}%
} \\
\makebox[\linewidth][c]{%
{\normalfont\small
\textsuperscript{2}Guangdong University of Technology}%
} \\
\makebox[\linewidth][c]{%
{\normalfont\small
\textsuperscript{\textdagger}Corresponding author.}%
}
}

\iclrfinalcopy % Uncomment for camera-ready version, but NOT for submission.
\begin{document}

\maketitle

\fancyhead{}
\renewcommand{\headrulewidth}{0pt}

\begin{abstract}
Recent task-driven and just-in-time 3D Scene Graph (3DSG) methods reduce per-task representations by constructing or activating only
task-relevant information. Yet sparse per-task working sets do not bound
onboard memory usage over a robot's lifetime: as tasks change, payloads
accumulated for earlier tasks may become irrelevant to the current task but can
be useful again in future tasks. Over repeated task switches and expanding environments, retaining such reusable payloads causes local
memory to grow, whereas discarding them entirely can lead to costly repeated
construction of the same payloads later.
We introduce \textbf{PORTER}, which decouples
persistence from residency: lightweight anchors remain in the limited memory
of the edge robot while heavy object payloads migrate between the edge and the
cloud. Relevance alone is insufficient for deciding residency because multiple relevant payloads may provide redundant information. We therefore decompose each task into functional requirements and
introduce \emph{Irreplaceable Support Erasure (ISE)}, which measures the loss
in requirement coverage caused by offloading. ISE discounts replaceable support and penalizes losses more strongly when
the remaining coverage of a requirement is weak. PORTER constructs a
budget-aware local working set by repeatedly offloading the payload with the
smallest marginal ISE per byte.
Experiments on JITOMA-Bench evaluate PORTER across four 3DSG builders. Under
progressive compression, pooled relative mR@3 remains at \textbf{100\%} through
\textbf{91\%} payload-byte offloading.
\end{abstract}

\section{Introduction}
\label{sec:introduction}

Long-lived embodied agents need memory that outlasts any individual task:
objects perceived today may become useful again minutes, days, or tasks later.
Yet retaining raw observations indefinitely is neither an efficient nor a
useful way to reason about a growing environment. 3D Scene Graphs (3DSGs)~\citep{gay2018visual, armeni20193d}
offer an appealing alternative by abstracting accumulated observations into
persistent, geometrically grounded objects and relations, yielding a compact
and queryable representation for downstream navigation, manipulation, and
scene reasoning~\citep{honerkamp2024language, yin2024sg, werby2024hierarchical, yan2025dynamic, chen2026psg, xia2026exploring}. This abstraction makes
3DSGs a natural substrate for long-term robot memory. But making the
representation persistent raises a different systems question: \emph{can the robot retain what it has learned without keeping all of it locally resident?}

Recent 3DSGs have become increasingly capable of serving reusable scene
memory. Open-vocabulary representations broaden what can be queried
~\citep{jatavallabhula2023conceptfusion, gu2024conceptgraphs,
koch2024open3dsg,werby2024hierarchical,yamazaki2024open,linok2025beyond,
fungraph3d,chang2026rag}, while task-driven and just-in-time methods avoid
processing the full scene for every goal by constructing, abstracting, or
activating only task-relevant information
~\citep{agia2022taskography,maggio2024clio,maggio2025bayesian,
maggio2026found,chang2026just}. These advances improve what the memory can
express and how efficiently it serves the \emph{current} task, but they do not
bound what accumulates over the robot's \emph{lifetime}. As tasks change,
payloads useful to earlier tasks may no longer deserve scarce local memory
while remaining valuable for future reuse. Over repeated task switches and
expanding environments, retaining such reusable payloads causes onboard memory
to grow, whereas deleting them turns efficiency into forgetting.
This exposes a missing distinction between \emph{persistence} and
\emph{residency}. An object need not disappear from the robot's memory merely
because its representation is no longer worth occupying the edge.

\begin{figure}[ht]
    \centering
    \includegraphics[width=\linewidth]{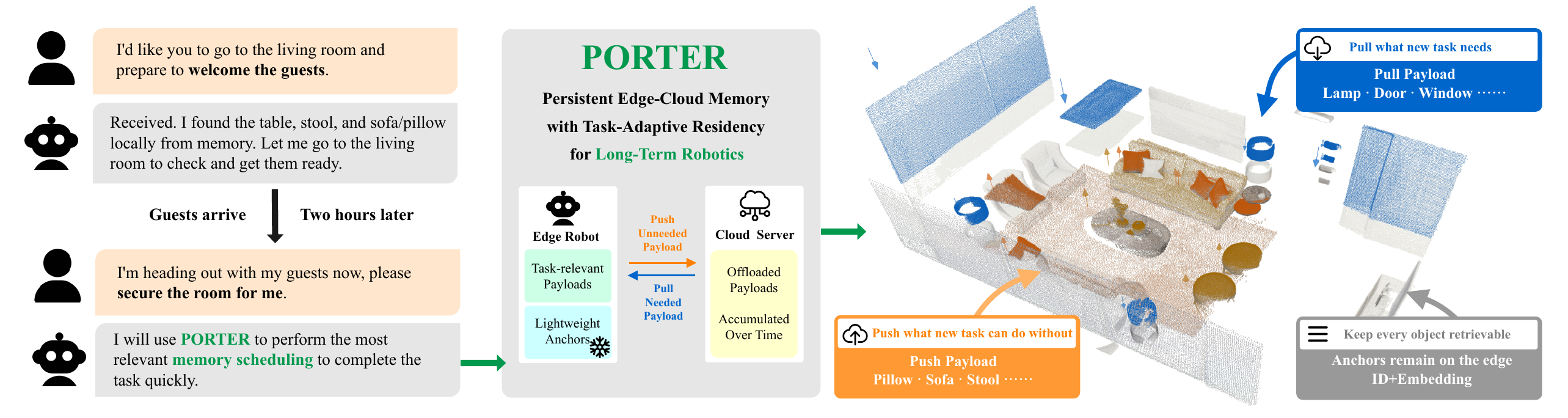}
    \caption{
\textbf{Overview of PORTER.}
PORTER keeps lightweight anchors on the robot while task-dependent payloads
move between the edge and cloud, allowing scene knowledge to persist across
tasks without requiring all payloads to remain locally resident.
}
    \label{fig:porter_overview}
\end{figure}

We therefore formulate \emph{task-switch-aware edge--cloud residency}, where scene
knowledge persists across tasks while its physical placement adapts to the task. \textbf{PORTER} realizes this idea by factorizing each 3DSG object into a
persistent lightweight anchor and a migratable heavy payload. The anchor
preserves identity and retrieval data locally, whereas the payload contains
the storage-intensive state supplied by the upstream 3DSG builder---for
example, dense object point clouds, historical image crops, or other
builder-specific observations. Because only the payload moves, PORTER can
release local memory without deleting the object, while remaining
agnostic to how that object was constructed. Task switching then
becomes a placement problem rather than a forgetting problem: which
edge-resident payloads should be pushed to the cloud, and which cloud-resident
payloads should be pulled back to form the working set for the new task?

Making this decision requires more than task relevance: what matters is how
much task support would be lost if a payload left \emph{now}. A chair and a
sofa may both satisfy the need for seating; either may incur little modeled support loss while the other remains, but the survivor becomes increasingly valuable once
its substitute is gone. PORTER models this state-dependent redundancy by
decomposing the task into functional requirements and aggregating alternatives
with a Noisy-OR model. We formalize the resulting counterfactual loss as
\emph{Irreplaceable Support Erasure (ISE)}, which accounts for both substitute
coverage and residual fragility. PORTER then constructs a budget-aware working
set by repeatedly offloading the payload with the smallest \emph{marginal ISE
per byte}, recomputing its value as the resident set changes.
Across four heterogeneous 3DSG builders on JITOMA-Bench~\citep{chang2026just},
PORTER reduces local residency while preserving task capability, with no mR@3
degradation through $91\%$ payload-byte offloading. On a real robot, it further
reduces average local residency by $59.1\%$ and mapping TPF by $70.3\%$
without reducing mR@3.
\section{Related Work}
\label{sec:related_work}

\subsection{3D Scene Graphs for Embodied Intelligence}

Scene graphs abstract visual environments into entities and relations, and
their extension to 3D yields geometrically grounded, object-centric scene
representations that are more structured and compact than dense sensory
observations
~\citep{johnson2015image,lu2016visual,krishna2017visual,armeni20193d}.
Early 3DSG systems further enabled online and hierarchical mapping for robotic
operation
~\citep{rosinol2021kimera,wu2021scenegraphfusion,hughes2022hydra}, while
recent vision--language models have extended 3DSGs beyond predefined semantic
categories to open-vocabulary representations
~\citep{jatavallabhula2023conceptfusion,gu2024conceptgraphs,
koch2024open3dsg,werby2024hierarchical,yamazaki2024open,linok2025beyond,
fungraph3d,chang2026rag}. These developments have made 3DSGs a versatile
interface for embodied tasks including navigation, object search,
manipulation, and high-level planning
~\citep{gadre2022clip,agia2022taskography,shah2023lm,rana2023sayplan,
yin2024sg,yan2025dynamic,chen2026psg,xia2026exploring}. More broadly, their
persistent object identities and relational structure allow accumulated scene
knowledge to be queried and reused across interactions, making 3DSGs a natural
substrate for embodied memory.

\subsection{Toward Persistent 3D Scene Graph Memory}

As open-vocabulary 3DSGs grow richer, maintaining all available scene
information conflicts with the resource constraints of embodied
platforms. Task-driven and just-in-time methods address this tension by selectively allocating
representation and computation to the current objective.
Taskography~\citep{agia2022taskography} introduced task-relevant reasoning over
large 3DSGs, while Clio~\citep{maggio2024clio} adapts open-set abstraction to
downstream tasks. Subsequent methods further explore task-conditioned
representation and granularity~\citep{maggio2025bayesian}, including
FOUND-IT~\citep{maggio2026found} and JITOMA~\citep{chang2026just}, which adapt
granularity or activate expensive semantics as task demands evolve.
Related systems treat 3DSGs as embodied memory:
GraphEQA~\citep{saxena2024grapheqa} combines 3DSGs with task-relevant
observations as multimodal memory, while KARMA~\citep{wang2025karma} integrates
them into a long- and short-term memory architecture. Yet selective abstraction
and activation alone do not bound long-term memory growth: as environments and
tasks evolve, reusable information continues to accumulate beyond what must
remain locally active. Recent analyses~\citep{rotondi20263d} accordingly
identify compact representation and memory management over months or years as
open challenges for long-term 3DSGs.
\section{Method}
\label{sec:method}

PORTER decouples logical \emph{persistence} from physical
\emph{residency}. Each 3DSG object retains a lightweight local anchor, while
its storage-intensive payload may migrate between the edge and the cloud.
The key question is not how relevant a payload is, but what task support would
be lost if it left the edge. We formalize this counterfactual loss as
\emph{Irreplaceable Support Erasure (ISE)}, which accounts for both substitute
support and the fragility of what remains. PORTER then constructs a
budget-aware working set by repeatedly offloading the payload with the smallest
\emph{marginal ISE per byte}, recomputing its value as the resident set changes.

\subsection{Persistent Anchors and Migratable Payloads}
\label{sec:method_memory}

Let $\mathcal V=\{v_1,\ldots,v_N\}$ denote the object nodes produced by an
upstream 3DSG. PORTER factorizes each object as
$\mathcal O_v=(A_v,P_v)$, where $A_v$ is a lightweight anchor that preserves
identity and compact retrieval data, including its vision--language
embedding for task-conditioned retrieval, and $P_v$ contains the
storage-intensive builder-specific state, such as a dense 3D point cloud,
historical object crops, or other per-object observations. Let $b_v$ denote the local memory footprint of $P_v$.

All anchors remain on the edge, while only payloads are subject to residency
decisions. Let $S\subseteq\mathcal V$ denote the objects whose payloads are
currently local, with memory footprint $B(S)=\sum_{v\in S}b_v$. Offloading
$v$ therefore moves only $P_v$ to the cloud while retaining $A_v$ locally, so
the object remains identifiable and retrievable in the persistent 3DSG.
Residency changes are thus reversible: a later task can restore $P_v$ without
reconstructing the object from scratch or revisiting the scene.

\subsection{From Relevance to Irreplaceable Task Support}
\label{sec:method_support}

Residency should depend not only on object relevance in isolation, but on
what task support would become unavailable without it. This
replaceability is \emph{requirement-dependent}: the same object may be
indispensable for one functional need yet redundant for another. 
% We therefore
% use an LLM to decompose task $q$ into a compact set of functional requirements
% $\mathcal R_q=\{r_1,\ldots,r_M\}$, with equal weights $\omega_r=1/M$.
We represent task $q$ as a compact set of functional requirements
$\mathcal R_q=\{r_1,\ldots,r_M\}$, with equal weights
$\omega_r=1/M$. The requirements may be supplied by a task parser or
language model; PORTER itself operates only on the
resulting requirement set.
This lets PORTER ask not only whether an object is related to the task, but
\emph{which requirement it supports and whether that support is already
provided elsewhere}.

We first estimate the potential support of object $v$ for requirement $r$.
Let $\hat{\mathbf f}_v$ and $\hat{\mathbf g}_r$ denote their normalized
embeddings in a shared vision--language space, with
$s_{vr}=\hat{\mathbf f}_v^{\mathsf T}\hat{\mathbf g}_r$. We define
\begin{equation}
    a^F_{vr}
    =
    \mathbf 1[v\in\mathcal K_r]\,
    \operatorname{clip}\!\left(
        \frac{s_{vr}-\alpha}{1-\alpha},
        0,\,
        1-\varepsilon
    \right),
    \qquad
    a^A_{vr}=\eta a^F_{vr},
    \label{eq:porter_support}
\end{equation}
where $\mathcal K_r$ contains the top-$K$ candidate objects for requirement
$r$, and $\eta\in[0,1]$ controls how much task support remains available from the
lightweight anchor after its payload is offloaded. Thus, $a^F_{vr}$ and
$a^A_{vr}$ represent the soft support associated with the full payload and
anchor-only states, respectively. These scores are not treated as calibrated
probabilities; their role is to expose how object support changes with
residency.
For a candidate local set $S$, the effective support becomes
\begin{equation}
    e_{vr}(S)=
    \begin{cases}
        a^F_{vr}, & v\in S,\\
        a^A_{vr}, & v\notin S.
    \end{cases}
    \label{eq:porter_effective_support}
\end{equation}
What matters, however, is not this support alone but how much of it remains
uncovered by alternatives. Under a Noisy-OR aggregation, $1-e_{vr}(S)$ is the
portion of requirement $r$ left unsupported by object $v$, and their product
therefore represents the support left uncovered by all objects. We define
\begin{equation}
    c_r(S)
    =
    1-\prod_{v\in\mathcal V}\bigl(1-e_{vr}(S)\bigr),
    \label{eq:porter_coverage}
\end{equation}
where $c_r(S)$ is the soft coverage of requirement $r$ under residency set
$S$.

This construction makes substitute redundancy explicit. When several objects
support the same requirement, its uncovered residual is already small and
another substitute contributes little additional coverage; as those
alternatives disappear, the remaining objects become progressively more
valuable. PORTER therefore does not equate relevance with value: an object's
value is the support that the rest of the current working set cannot replace.
The next section measures the erasure of precisely this irreplaceable support under payload offloading, turning replaceability into a residency criterion.

\subsection{Irreplaceable Support Erasure}
\label{sec:method_ise}

The coverage $c_r(S)$ above tells us what requirement support remains after
accounting for substitutes. Residency, however, requires the complementary
quantity: how much support is erased as payloads leave the edge. Importantly,
this loss should be measured against what the upstream 3DSG could provide
under full residency, so that PORTER is charged only for information lost
through offloading rather than support that was never available in the first
place.

For requirement $r$, we therefore define the \emph{support-survival ratio}
\begin{equation}
    \rho_r(S)
    =
    \frac{c_r(S)+\epsilon_{\mathrm{ISE}}}
         {c_r(\mathcal V)+\epsilon_{\mathrm{ISE}}},
    \qquad
    0<\rho_r(S)\leq 1,
    \label{eq:porter_survival}
\end{equation}
where $\epsilon_{\mathrm{ISE}}>0$ is a small numerical stabilizer. Since
$a^A_{vr}\leq a^F_{vr}$, full residency provides maximal support and
$\rho_r(\mathcal V)=1$; smaller values indicate that more of the originally
available support for requirement $r$ has been erased locally.

A linear support drop, however, does not distinguish whether the remaining
requirement is still well covered or close to losing its support entirely.
We therefore calibrate relative erasure through a negative logarithm and define \emph{Irreplaceable Support Erasure (ISE)} as
\begin{equation}
    \operatorname{ISE}_q(S)
    =
    -\sum_{r\in\mathcal R_q}
    \omega_r\log_2 \rho_r(S)
    =
    \sum_{r\in\mathcal R_q}
    \omega_r
    \log_2
    \frac{c_r(\mathcal V)+\epsilon_{\mathrm{ISE}}}
         {c_r(S)+\epsilon_{\mathrm{ISE}}}.
    \label{eq:porter_ise}
\end{equation}
Thus, $\operatorname{ISE}_q(\mathcal V)=0$, while the same absolute support
drop incurs a larger penalty when the remaining coverage is already weak.
For example, ignoring $\epsilon_{\mathrm{ISE}}$ for illustration, reducing a
requirement from $0.9$ to $0.8$ contributes only
$\log_2(0.9/0.8)\approx0.17$, whereas reducing it from $0.2$ to $0.1$
contributes $1$ bit before weighting. The logarithm therefore protects
requirements that are close to losing their remaining support, rather than
treating all coverage drops equally. Because $c_r(S)$ is soft
vision--language evidence rather than a calibrated probability, we use the
logarithm as an information-style calibration of relative erasure rather than
a Shannon entropy estimate.

The target residency set can then be defined by minimizing ISE under the local
memory budget,
\begin{equation}
    S_q^\star
    \in
    \arg\min_{S\subseteq\mathcal V}
    \operatorname{ISE}_q(S)
    \qquad
    \mathrm{s.t.}\quad
    B(S)\leq B.
    \label{eq:porter_objective}
\end{equation}
ISE is the complement of a monotone submodular retained-support utility,
making Eq.~\ref{eq:porter_objective} a budgeted submodular maximization problem
(Appendix~\ref{app:ise_derivation}).

To turn this set-level objective into a payload-level decision, consider
offloading $P_v$ for $v\in S$. The Noisy-OR construction gives the resulting
drop in requirement $r$ exactly as
\begin{equation}
    \delta_{vr}(S)
    :=
    c_r(S)-c_r(S\setminus\{v\})
    =
    \bigl(a^F_{vr}-a^A_{vr}\bigr)
    \prod_{u\neq v}\bigl(1-e_{ur}(S)\bigr).
    \label{eq:porter_support_drop}
\end{equation}
This factorization exposes replaceability: the first term is the
support removed with $P_v$, while the product measures how much of that support
is not already covered by the remaining objects.

Substituting this counterfactual drop into Eq.~\ref{eq:porter_ise} yields the
\emph{marginal ISE} of offloading $P_v$,
\begin{equation}
    \boxed{
    \Delta_q^{\mathrm{ISE}}(v;S)
    :=
    \operatorname{ISE}_q(S\setminus\{v\})
    -
    \operatorname{ISE}_q(S)
    =
    \sum_{r\in\mathcal R_q}
    \omega_r
    \log_2\!\left(
        \frac{c_r(S)+\epsilon_{\mathrm{ISE}}}
             {c_r(S)-\delta_{vr}(S)+\epsilon_{\mathrm{ISE}}}
    \right)
    }.
    \label{eq:porter_marginal}
\end{equation}
Notably, the full-residency reference cancels in the marginal, leaving a purely state-dependent ratio between the requirement coverage before and after offloading \(P_v\).
Marginal ISE therefore captures two complementary effects: the
counterfactual drop $\delta_{vr}(S)$ discounts support that can be replaced by
substitutes, while the logarithmic ratio amplifies losses to requirements whose
remaining coverage is already fragile. It asks not ``how relevant is $v$?'',
but \emph{what would the task lose if $P_v$ were offloaded?}

Crucially, this value depends on the current residency set $S$. A payload that
is redundant now may become irreplaceable after its substitutes are offloaded:
both its uncovered support and the fragility of the corresponding requirement
increase as the working set shrinks. Marginal ISE must therefore be recomputed
as residency changes, which motivates the state-dependent scheduler introduced
next.

\subsection{ISE-Aware Residency Selection}
\label{sec:method_scheduler}

We operationalize this objective with a lightweight reverse-greedy removal rule. At each
step, PORTER offloads the resident payload whose marginal ISE per freed byte is
smallest,
\begin{equation}
    v^\star
    =
    \arg\min_{v\in S}
    \frac{\Delta_q^{\mathrm{ISE}}(v;S)}{b_v}.
    \label{eq:porter_reverse_greedy}
\end{equation}
Starting from full residency $S_0=\mathcal V$, reverse greedy repeats this
decision until the local budget $B(S)\leq B$ is satisfied, yielding the target
residency set $\widehat S_q$. Crucially, $\Delta_q^{\mathrm{ISE}}(v;S)$ is
recomputed after every offload: a payload that is redundant early can become
costly to remove once its substitutes disappear. PORTER therefore allocates
scarce local memory to support that is hardest to erase safely, rather than
simply retaining the objects most relevant to the current task.

\subsection{Residency Across Task Switches}
\label{sec:method_reconcile}

The target $\widehat S_q$ depends on the current task, but not on the residency
left by previous tasks; the latter determines only which transfers are needed.
Given the current local set $\mathcal L_t$, PORTER realizes the new target by
reconciling the two residency sets,
\begin{equation}
    \mathrm{Push}_t
    =
    \mathcal L_t\setminus\widehat S_q,
    \qquad
    \mathrm{Pull}_t
    =
    \widehat S_q\setminus\mathcal L_t.
    \label{eq:porter_reconciliation}
\end{equation}
Push sends payloads that no longer belong in the target working set to the
cloud, while Pull restores newly required payloads to the edge. Because their
anchors persist locally, both operations preserve object identity without
modifying the upstream 3DSG or reconstructing the scene representation.
In the current implementation, a pushed payload is frozen while cloud-resident;
Pull restores its latest persisted state, after which the upstream builder
resumes native updates.

% Together, these components close the PORTER loop: Noisy-OR models what support
% can be replaced, ISE measures what would be erased, marginal ISE per byte
% decides what deserves local residency, and Push/Pull operations realize that
% target as tasks continue to change.
\section{Experiments}
\label{sec:experiments}

We evaluate whether PORTER can reduce local 3DSG residency while preserving
task capability under changing task demands. Our experiments address two
questions:
\textbf{(Q1)} Can PORTER act as a construction-agnostic residency layer across
heterogeneous 3DSG builders without materially compromising their task
capability?
\textbf{(Q2)} Does PORTER preserve task capability more effectively than
simpler offloading policies as local scene memory is progressively compressed?

\subsection{Experimental Setup}
\label{sec:exp_setup}

\paragraph{Benchmark and protocol.}
JITOMA-Bench~\citep{chang2026just}, built upon the real-world data of
Clio-Bench~\citep{maggio2024clio}, contains three scenes with 35 Tier~2 and 55 Tier~3
ground-truth target cases. We use two
complementary evaluation tracks. Tier~2 executes a sequence of tasks over the same persistent scene memory: after one task is evaluated, the next begins without resetting the accumulated representation. 
Tier~3 instead contains complex instructions
involving multiple primary and latent targets, providing a complementary
setting for testing how well a residency policy preserves multi-requirement
task capability as local scene memory is progressively compressed.

\paragraph{Evaluation metrics and control.}
For Tier~2, we follow JITOMA-Bench and report top-1 \textbf{IoU},
\textbf{mR@1}, \textbf{mR@3}, \textbf{Objs}, \textbf{Peak}, and
\textbf{TPF}. mRecall averages recall over IoU thresholds
$\{0.1,0.2,0.3\}$; Objs and Peak are the average and maximum numbers of
edge-resident objects, and TPF is the average time per frame.
For Tier~3, we report \textbf{relative mR@3}, normalized to the
Keep-All reference, against cumulative payload bytes offloaded.
All methods share the same RGB-D trajectories, task annotations, and ground truth; PORTER changes only payload residency.
Tier~3 policies additionally begin each task from the identical Keep-All
ConceptGraphs state and differ only in payload-removal order.
Implementation details and hyperparameters are provided in
Appendix~\ref{app:implementation}.

\subsection{Construction-Agnostic Residency}
\label{sec:exp_mapper_agnostic}

To answer \textbf{Q1}, we deliberately evaluate PORTER on four 3DSG builders
with substantially different construction philosophies:
ConceptGraphs~\citep{gu2024conceptgraphs}, a bottom-up open-vocabulary mapping
pipeline; ReasoningGraph~\citep{puigjaner2026relationship}, which emphasizes
hierarchical object--relational reasoning; Clio~\citep{maggio2024clio}, a
real-time task-driven scene graph; and DAAAM~\citep{gorlo2026describe}, which
constructs hierarchical 4D spatio-temporal memory. Together, these methods span
distinct design choices in when, how, and at what structure scene information
is instantiated, providing a diverse testbed for PORTER's construction-agnostic
design.

For each builder, we compare its native execution with the same method augmented
by PORTER, while keeping its original construction procedure and parameters
unchanged. The paired variants differ only in payload residency across task
switches: PORTER changes which payload-bearing objects remain locally available
for subsequent updates, while offloaded objects persist through their
lightweight anchors and can be restored when needed. This paired design tests
whether the same residency principle can reduce the local working
representation across heterogeneous 3DSG paradigms without materially
compromising the task capability inherited from each builder.
Table~\ref{tab:porter_method_agnostic} reports the resulting
task-performance--efficiency trade-off.

% ==========================================
% PORTER: Method-Agnostic Evaluation
% JITOMA-Bench Tier 2 only
%
% Requires:
% \usepackage{xcolor,colortbl,multirow,booktabs,graphicx}
% ==========================================

\definecolor{accBg}{HTML}{FFF0F5} % Accuracy
\definecolor{effBg}{HTML}{FFFFF0} % Efficiency
\definecolor{porterBg}{HTML}{EEF8EE} % PORTER row

\begin{table*}[ht]
\centering
\renewcommand{\arraystretch}{1.15}
\setlength{\tabcolsep}{4.5pt}

\caption{\textbf{Construction-agnostic evaluation on JITOMA-Bench \citep{chang2026just} Tier~2.}
We pair four heterogeneous 3DSG builders with PORTER while keeping their
native construction pipelines unchanged. For each scene, we report 
grounding accuracy (\textbf{IoU}, \textbf{mR@1}, and \textbf{mR@3}) and
system efficiency (\textbf{Objs}, \textbf{Peak}, and \textbf{TPF}).
The paired comparison tests whether PORTER can reduce the
local payloads across different 3DSG paradigms
while largely preserving the task capability.}
\label{tab:porter_method_agnostic}

\resizebox{\textwidth}{!}{%
\begin{tabular}{
l l
>{\columncolor{accBg}}c >{\columncolor{accBg}}c >{\columncolor{accBg}}c
>{\columncolor{effBg}}c >{\columncolor{effBg}}c >{\columncolor{effBg}}c
>{\columncolor{accBg}}c >{\columncolor{accBg}}c >{\columncolor{accBg}}c
>{\columncolor{effBg}}c >{\columncolor{effBg}}c >{\columncolor{effBg}}c
>{\columncolor{accBg}}c >{\columncolor{accBg}}c >{\columncolor{accBg}}c
>{\columncolor{effBg}}c >{\columncolor{effBg}}c >{\columncolor{effBg}}c
}

\toprule

\multirow{3}{*}{\textbf{3DSG Builder}}
&
\multirow{3}{*}{\textbf{Method}}
&
\multicolumn{6}{c}{\textbf{Apartment}}
&
\multicolumn{6}{c}{\textbf{Office}}
&
\multicolumn{6}{c}{\textbf{Cubicle}}
\\

\cmidrule(lr){3-8}
\cmidrule(lr){9-14}
\cmidrule(lr){15-20}

&
&
\multicolumn{3}{c}{\cellcolor{accBg}\textit{Accuracy $\uparrow$}}
&
\multicolumn{3}{c}{\cellcolor{effBg}\textit{Efficiency}}
&
\multicolumn{3}{c}{\cellcolor{accBg}\textit{Accuracy $\uparrow$}}
&
\multicolumn{3}{c}{\cellcolor{effBg}\textit{Efficiency}}
&
\multicolumn{3}{c}{\cellcolor{accBg}\textit{Accuracy $\uparrow$}}
&
\multicolumn{3}{c}{\cellcolor{effBg}\textit{Efficiency}}
\\

\cmidrule(lr){3-5}
\cmidrule(lr){6-8}
\cmidrule(lr){9-11}
\cmidrule(lr){12-14}
\cmidrule(lr){15-17}
\cmidrule(lr){18-20}

&
&
IoU & mR@1 & mR@3
& Objs $\downarrow$ & Peak $\downarrow$ & TPF $\downarrow$
&
IoU & mR@1 & mR@3
& Objs $\downarrow$ & Peak $\downarrow$ & TPF $\downarrow$
&
IoU & mR@1 & mR@3
& Objs $\downarrow$ & Peak $\downarrow$ & TPF $\downarrow$
\\

\midrule

% =========================================================
% ConceptGraphs
% =========================================================
\multirow{2}{*}{ConceptGraphs}
& Vanilla
& 17.8 & 33.3 & 52.8
& 452 & 1476 & 6.03
& 19.0 & 40.5 & 45.2
& 924 & 5796 & 12.58
& 11.3 & 22.2 & 63.0
& 224 & 745 & 5.37
\\

& \cellcolor{porterBg}\textbf{+ PORTER}
& 17.6 & 33.3 & 55.6
& 83 & 341 & 3.96
& 19.2 & 40.5 & 45.2
& 128 & 1726 & 5.02
& 12.4 & 22.2 & 63.0
& 75 & 270 & 3.98
\\

\midrule

% =========================================================
% ReasoningGraph
% =========================================================
\multirow{2}{*}{ReasoningGraph}
& Vanilla
& 12.6 & 33.3 & 33.3
& 276 & 488 & 2.91
& 17.3 & 42.9 & 45.2
& 261 & 605 & 3.05
& 12.9 & 22.2 & 29.6
& 149 & 260 & 2.92
\\

& \cellcolor{porterBg}\textbf{+ PORTER}
& 12.6 & 33.3 & 33.3
& 5 & 92 & 2.91
& 17.3 & 42.9 & 45.2
& 4 & 83 & 3.05
& 12.9 & 22.2 & 29.6
& 4 & 59 & 2.92
\\

\midrule

% =========================================================
% Clio
% =========================================================
\multirow{2}{*}{Clio}
& Vanilla
& 10.9 & 25.0 & 27.8
& 4 & 637 & 0.35
& 17.7 & 38.1 & 47.6
& 5 & 678 & 0.42
& 23.8 & 66.7 & 74.1
& 5 & 408 & 0.46
\\

& \cellcolor{porterBg}\textbf{+ PORTER}
& 10.9 & 25.0 & 27.8
& 3 & 74 & 0.38
& 17.9 & 40.5 & 47.6
& 3 & 65 & 0.40
& 24.5 & 66.7 & 74.1
& 5 & 75 & 0.41
\\

\midrule

% =========================================================
% DAAAM
% =========================================================
\multirow{2}{*}{DAAAM}
& Vanilla
& 11.6 & 22.2 & 33.3
& 739 & 1327 & 0.16
& 14.2 & 33.3 & 50.0
& 713 & 1738 & 0.14
& 12.6 & 29.6 & 48.2
& 417 & 710 & 0.15
\\

& \cellcolor{porterBg}\textbf{+ PORTER}
& 13.9 & 27.8 & 33.3
& 61 & 745 & 0.16
& 14.9 & 35.7 & 46.7
& 134 & 602 & 0.14
& 14.2 & 33.3 & 51.9
& 196 & 442 & 0.15
\\

\bottomrule
\end{tabular}
}
\end{table*}

\paragraph{Results.}
Table~\ref{tab:porter_method_agnostic} shows that PORTER consistently reduces
local residency across all four 3DSG builders while preserving task capability.
Across ConceptGraphs, ReasoningGraph, Clio, and DAAAM, PORTER reduces the
average resident object count by $82.1\%$, $98.1\%$, $21.4\%$, and $79.1\%$,
respectively. These results support PORTER's use as a
construction-agnostic residency layer over heterogeneous 3DSGs under a
unified task-conditioned residency objective. 
Accuracy remains broadly comparable across IoU, mR@1, and mR@3, with several
entries even improving. In some cases, reducing the resident candidate set can
suppress distracting evidence while preserving useful retrievals, as
illustrated by the Clio case studies in
Appendix~\ref{app:case_study}.
This suggests that ISE preserves irreplaceable task support while
removing redundant payloads.

Beyond memory reduction, PORTER improves online efficiency when computation
scales with resident objects. For ConceptGraphs, PORTER reduces average TPF
from 6.03s to 3.96s in Apartment, 12.58s to 5.02s in Office, and 5.37s to
3.98s in Cubicle. For builders with weaker runtime dependence on residency,
such as Clio and DAAAM, PORTER primarily provides memory reduction while maintaining similar execution time without modifying their native
construction pipelines.

\subsection{Residency under Progressive Compression}
\label{sec:exp_scheduler}

To answer \textbf{Q2}, we fix the upstream representation to the complete
ConceptGraphs~\citep{gu2024conceptgraphs} reconstruction and use the Tier~3
tasks of JITOMA-Bench~\citep{chang2026just}. Their multiple primary and latent
targets provide a discriminative setting for testing whether an offloading
policy preserves the different functional requirements of a task rather than
only its most salient object. Using a fixed, task-agnostic 3DSG also isolates
the residency decision from any upstream construction.

For every Tier~3 task, all policies start from the same Keep-All state. Each
policy then offloads one object payload at a time according to its own ranking
rule. After every offload, the task is re-evaluated using only the
payload-bearing objects that remain locally resident; offloaded payloads do not
participate in that evaluation. 
At each cumulative payload-byte checkpoint, we pool mR@3 across all Tier~3
tasks and normalize it by the pooled Keep-All mR@3 over the same task set.
% We record the cumulative fraction of payload
% bytes offloaded and normalize the resulting mR@3 by the task's Keep-All mR@3.
% Curves are then averaged over all Tier~3 tasks across the three scenes. 
The resulting trajectory measures how gracefully each policy preserves task
capability as increasingly more of the local 3DSG is moved to the cloud.

\paragraph{Compared policies.}
We compare five offloading strategies.
\textbf{Random} removes payloads in random order and provides a task-agnostic
reference.
\textbf{Largest-First} greedily removes the largest remaining payload,
capturing a purely memory-driven strategy.
\textbf{CLIP/Byte} combines direct task--object semantic relevance with payload
size.
\textbf{Requirement/Byte} decomposes the task into the same functional
requirements used by PORTER and scores objects from their requirement support,
but does not model substitute coverage.
\textbf{PORTER} instead ranks each candidate by its current marginal ISE per
byte and recomputes this quantity after every removal. Additional comparisons
with redundancy-aware baselines are provided in Appendix~\ref{app:strong_baselines}.

% These policies represent increasingly informed notions of payload value:
% Random uses no task signal, Largest-First uses memory cost alone, CLIP/Byte adds
% direct task relevance, and Requirement/Byte adds functional task decomposition.
% PORTER additionally asks which of that support would actually become
% unavailable under the \emph{current} residency state.

\begin{figure}[ht]
    \centering
    \includegraphics[width=\linewidth]{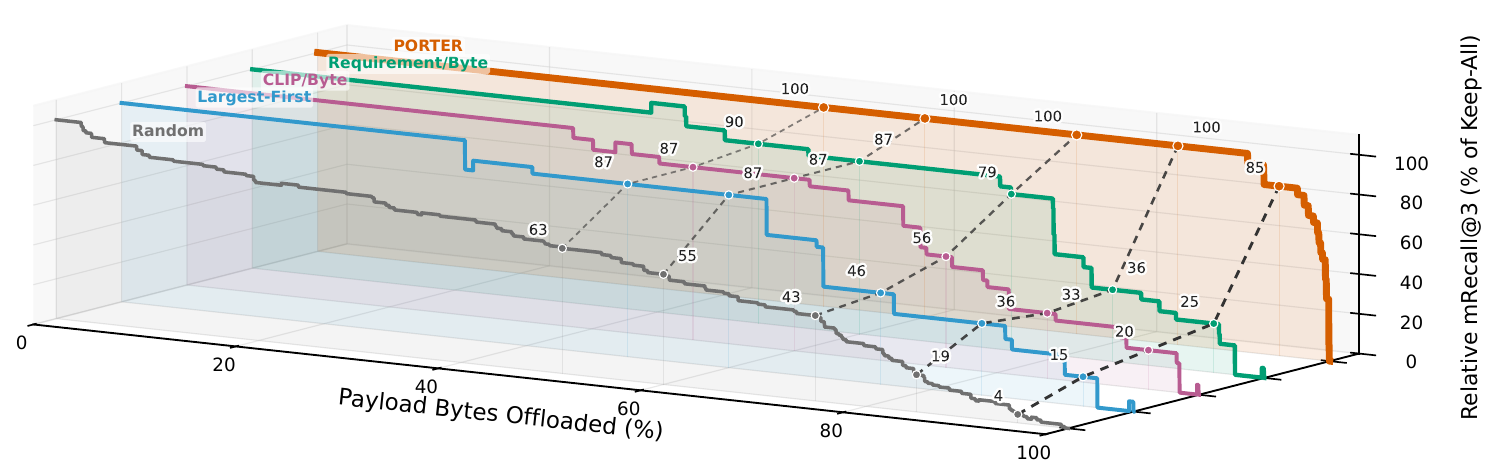}
    \caption{
    \textbf{Task retention under progressive 3DSG compression.}
    All policies start each Tier~3 task from the same ConceptGraphs
    representation and sequentially offload object payloads. The horizontal
    axis reports cumulative payload bytes offloaded, while the vertical axis
    reports mR@3 relative to Keep-All. Methods are separated along the depth
    axis only for visualization; dashed lines compare policies at matched
    offloading ratios of $50\%$, $60\%$, $75\%$, $85\%$, and $95\%$.
    }
    \label{fig:porter_policy_curve}
\end{figure}

\paragraph{Results.}
Figure~\ref{fig:porter_policy_curve} reveals a separation between
relevance-based selection and state-dependent residency. PORTER retains $100\%$ of Keep-All mR@3 through the $85\%$ plotted checkpoint
($91\%$ on the full trajectory),
whereas the strongest competing policy in Figure~\ref{fig:porter_policy_curve} retains only $90\%$, $87\%$, $79\%$,
and $36\%$ at the matched $50\%$, $60\%$, $75\%$, and $85\%$ checkpoints,
respectively. At $95\%$ offloading, PORTER still preserves $85\%$ of the
Keep-All capability, while the baseline retains only $25\%$.

The baseline ordering clarifies where the gain comes from. Random and
Largest-First ignore task value; CLIP/Byte adds semantic relevance, and
Requirement/Byte further resolves functional requirements, yet both degrade
once compression becomes aggressive. Figure~\ref{fig:porter_policy_curve}
therefore focuses on this diagnostic progression, while stronger
redundancy-aware set-selection baselines are evaluated in
Appendix~\ref{app:strong_baselines}. These baselines remain competitive under
moderate compression but fall behind PORTER in the high-compression regime,
consistent with the view that redundancy-aware coverage explains much of the early
robustness, while ISE's fragility-aware protection becomes important as the
resident set approaches the task-conditioned working set.

\subsection{Real-Robot Deployment}
\label{sec:real_robot}

\begin{figure}[ht]
    \centering
    \includegraphics[width=\linewidth]{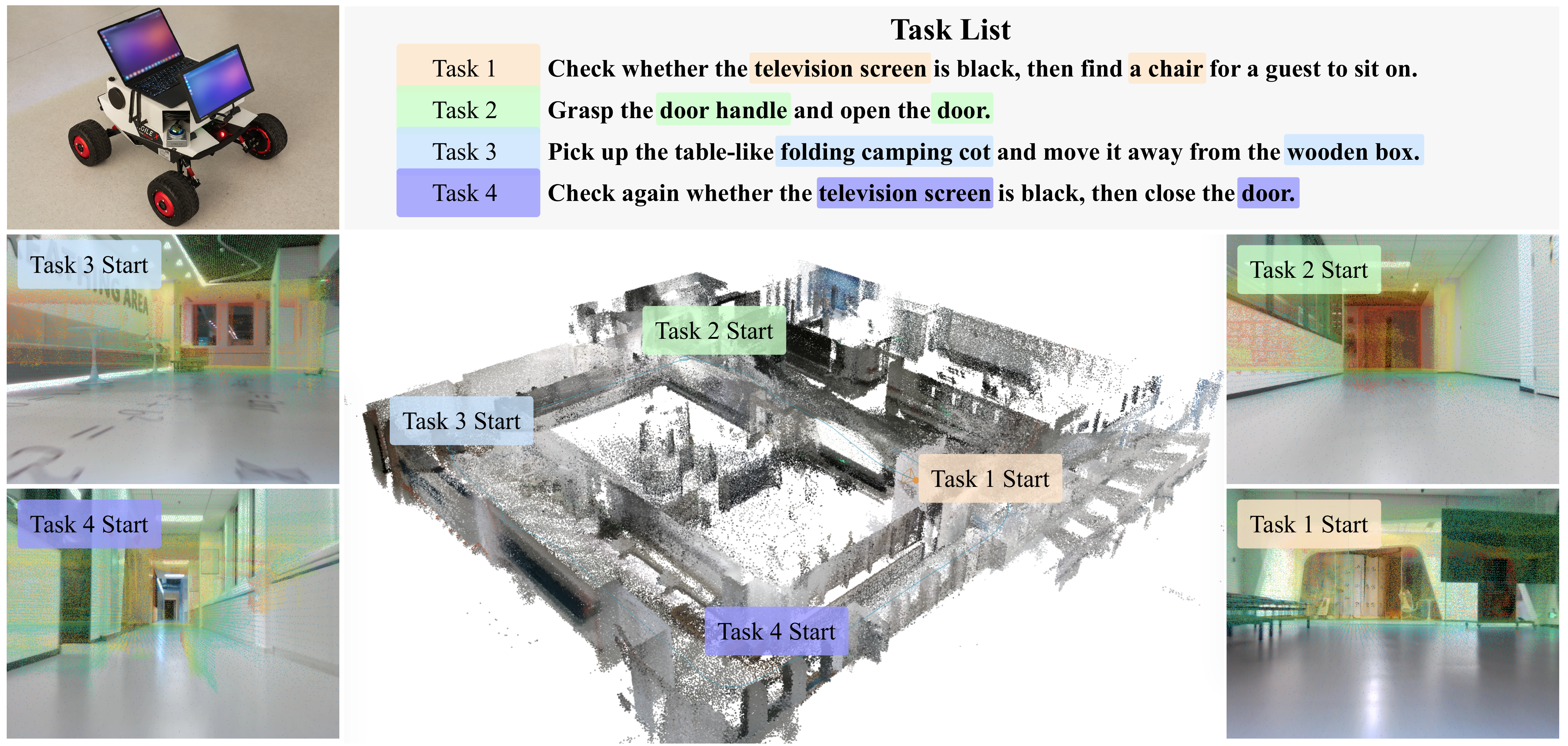}
    \caption{
    \textbf{Real-robot deployment.}
    The robot continuously maps an indoor environment while executing four
    consecutive tasks. The central reconstruction shows the task
    start locations; surrounding views show the corresponding observations.
    Repeated requirements, such as the television and door, test whether
    offloaded scene knowledge can later be restored when it becomes useful
    again.
    }
    \label{fig:real_robot_setup}
\end{figure}

We further evaluate PORTER on a real mobile robot under continuous scene
construction and task switching. The platform uses an Intel RealSense D435
RGB-D camera and two Livox Mid-360 LiDARs, with FAST-LIO2~\citep{xu2022fast}
providing odometry. Registered LiDAR geometry is reprojected into the camera
views to refine depth, and the resulting RGB-D sequence is used to construct
ConceptGraphs and annotate task-relevant 3D ground truth. We execute four
consecutive tasks over a 1,183-frame indoor trajectory, including repeated
requirements that test whether previously offloaded knowledge can be restored
(Figure~\ref{fig:real_robot_setup}). We report mR@3 using the benchmark IoU
thresholds; additional sensing, annotation, and communication details are
provided in Appendix~\ref{app:real_robot_details}.

\begin{wrapfigure}{r}{0.5\textwidth}
    \vspace{-0.7\baselineskip}
    \centering
    \includegraphics[width=\linewidth]{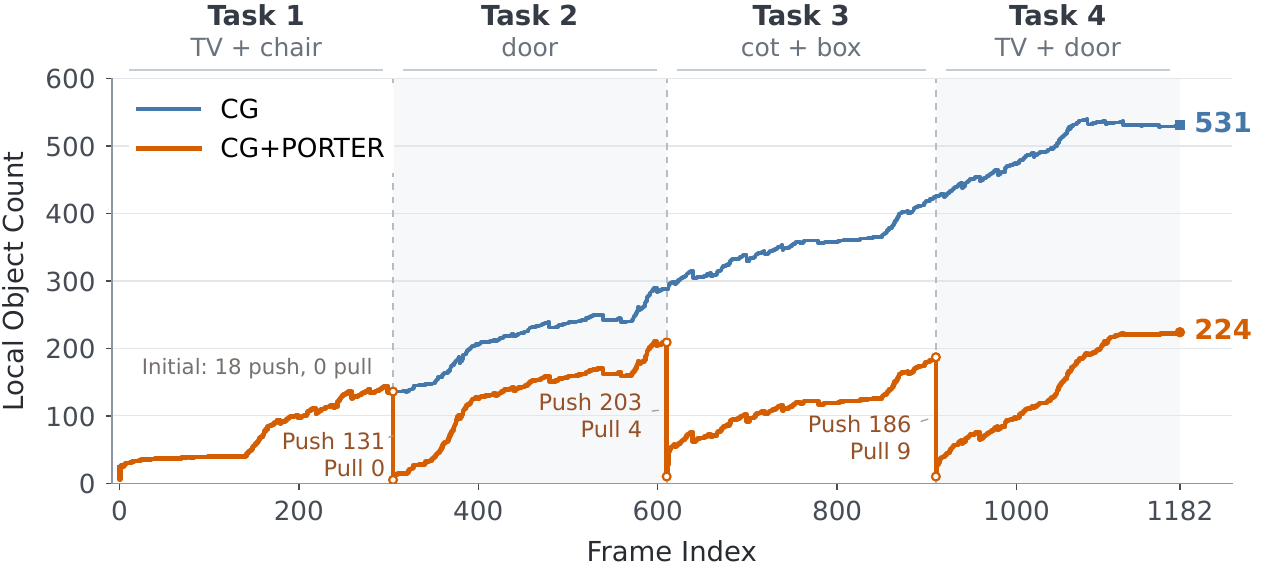}
    \caption{
    \textbf{Real-robot residency.}
    Vanilla ConceptGraphs continually accumulates objects, whereas PORTER
    contracts the working set at each task switch and resumes mapping from the new
    task-conditioned residency state.
    }
    \label{fig:real_robot_residency}
    \vspace{-0.5\baselineskip}
\end{wrapfigure}

Compared with vanilla ConceptGraphs, PORTER reduces the average number of
locally resident objects from $278$ to $114$
($59.1\%$) and the peak from $540$ to $224$ ($58.5\%$), while preserving
the same mR@3 of $75.0\%$. The smaller local working set also reduces mapping
time per frame from $6.276$\,s to $1.863$\,s ($70.3\%$).
Figure~\ref{fig:real_robot_residency} shows the mechanism behind this gain:
each task switch sharply removes payloads that are no longer needed, while
continued mapping subsequently grows the local set again. Crucially, the
later switches also restore previously offloaded objects---T3 and T4 trigger
4 and 9 Pulls, respectively---showing that PORTER changes \emph{residency}
rather than irreversibly forgetting scene knowledge.

\subsection{Component Ablations}
\label{sec:exp_ablation}

Having established PORTER's effectiveness in both controlled benchmarks and
real-robot deployment, we next isolate the design choices responsible for its advantage. We follow the same Tier~3
progressive-compression setting as Sec.~\ref{sec:exp_scheduler}, changing one
component at a time.
\textbf{w/o Noisy-OR} replaces Noisy-OR coverage with maximum support;
\textbf{w/o Log-Information Loss} replaces negative-log survival with linear
coverage loss;
\textbf{w/o Byte Normalization} ranks payloads by marginal ISE alone; and
\textbf{w/o Dynamic Re-ranking} freezes priorities computed from the Keep-All
state.

\begin{table*}[ht]
    \centering
    \small
    \caption{
\textbf{Component ablation of PORTER on JITOMA-Bench Tier~3.}
Overall nAUC covers the full payload-byte offloading curve and Tail nAUC
the high-compression interval $[95\%,100\%]$; remaining entries report pooled mR@3 relative to the pooled Keep-All reference.
}
    \label{tab:porter_component_ablation}
    \setlength{\tabcolsep}{0pt}
    \begin{tabular*}{\textwidth}{
        @{\extracolsep{\fill}}lccrrrrr@{}
    }
        \toprule
        & \multicolumn{2}{c}{Curve nAUC (\%) $\uparrow$}
        & \multicolumn{5}{c}{
            Relative $\mathrm{mRecall@3}$ at $\rho_B$ (\%) $\uparrow$
        } \\
        \cmidrule(lr){2-3}
        \cmidrule(lr){4-8}
        Variant
        & Overall
        & Tail
        & $95\%$
        & $97\%$
        & $99\%$
        & $99.5\%$
        & $99.9\%$ \\
        \midrule

        \textsc{Porter} w/o Noisy-OR
        & 61.20 & 35.93
        & 39.34 & 39.34 & 36.07 & 26.23 & 0.00 \\

        \textsc{Porter} w/o Byte Normalization
        & 97.89 & 62.42
        & 80.33 & 72.13 & 45.90 & 16.39 & 0.00 \\

        \textsc{Porter} w/o Log-Information Loss
        & 97.95 & 66.55
        & 81.97 & 72.13 & 50.82 & 40.98 & 19.67 \\

        \textsc{Porter} w/o Dynamic Re-ranking
        & 98.27 & 69.51
        & 85.25 & 77.05 & 54.10 & 40.98 & 19.67 \\

        \midrule

        \textbf{\textsc{Porter}}
        & \textbf{98.30}
        & \textbf{72.03}
        & \textbf{85.25}
        & \textbf{81.97}
        & \textbf{59.02}
        & \textbf{50.82}
        & \textbf{31.15} \\

        \bottomrule
    \end{tabular*}
\end{table*}

\paragraph{Results.}
Table~\ref{tab:porter_component_ablation} shows that full PORTER achieves the
best Overall and Tail nAUC ($98.30\%$ and $72.03\%$), with the largest drop
caused by removing Noisy-OR ($61.20\%$ and $35.93\%$). Most other differences
emerge only under high compression, when irrelevant and redundant payloads
have largely been removed and the resident set approaches the task-relevant
working set.

The tail results expose complementary roles of PORTER's components: Noisy-OR
captures replaceability, dynamic re-ranking updates it as substitutes
disappear, log-survival protects fragile requirements, and byte normalization
balances task loss against memory released. At $99.9\%$ offloading, PORTER
retains $31.15\%$ of Keep-All mR@3, versus $19.67\%$ without dynamic
re-ranking or log-information loss, while the no-byte and no-Noisy-OR variants
reach zero. We treat this tail as a stress test rather than a nominal operating
point. Parameter sensitivity is reported in
Appendix~\ref{app:parameter_sensitivity}.

\subsection{Runtime and Online Potential}
\label{sec:exp_runtime}

Our current implementation recomputes PORTER only at task switches. Across
25 Tier~3 cases, the full residency core takes only $2.034$\,ms median
(P95: $3.938$\,ms) on one pinned AMD EPYC~7542 logical CPU. Together with
the linear complexity derived in Appendix~\ref{app:porter_complexity}, this
suggests that ISE can support more frequent or asynchronous residency updates
as task context, memory pressure, or scene state evolves. Detailed timings and scene-size scaling are reported in
Appendix~\ref{app:runtime_details}.
These timings exclude physical payload transfer. In our real-robot deployment,
bulk Push transfers run asynchronously with mapping, while Pull is synchronous
because newly required payloads must be restored before use. As illustrated in
Figure~\ref{fig:real_robot_residency}, the task-critical Pull path is sparse:
T3 and T4 restore $3.72$\,MB and $19.23$\,MB in $0.76$\,s and
$1.24$\,s, respectively, while the substantially larger Push transfers run
concurrently with continued mapping. Detailed transfer measurements and connectivity
considerations are provided in Appendix~\ref{app:real_robot_transfer}.
\section{Conclusion}
\label{sec:conclusion}

We introduced PORTER, a construction-agnostic edge--cloud residency layer for
persistent 3DSG memory that decouples persistence from local
residency. Its Irreplaceable Support Erasure (ISE) objective measures task-support loss by accounting for substitute support and
residual fragility. Across heterogeneous 3DSG builders, progressive compression,
and real-robot deployment, PORTER reduces local residency while preserving task
capability, with additional mapping savings when computation scales with the
resident set. 
Its millisecond-scale decision cost further supports ISE
as a lightweight criterion for memory adaptation.
Overall, our results suggest that persistent scene knowledge can remain recoverable
without remaining continuously resident on the edge.

% 注意要去掉clearpage
\clearpage

\subsection*{AI use statement}

In this work, we did not use generative AI tools for research ideation or generating synthetic data.
% We used generative AI tools to aid and polish the writing of the manuscript,
% including improving clarity, readability, and presentation, and to proofread
% author-derived mathematical notation for possible omissions, inconsistencies,
% or simple algebraic errors. This mathematical proofreading did not involve
% generating mathematical claims.
We used generative AI tools to assist with writing, presentation, and
proofreading mathematical derivations for possible omissions, inconsistencies,
or algebraic errors. All mathematical statements and derivations in the paper
were independently checked and verified by the authors.
Generative AI was not used to generate figures
or experimental results. We take responsibility for the final content of this
work, including text or claims produced with the aid of generative AI.

% \subsection*{Ethics statement}

% (This section is \textbf{recommended} and does not count toward the page limit.)

% If authors feel that their paper submission raises questions regarding the Code
% of Ethics, they are encouraged to include a paragraph of Ethics Statement (at
% the end of the main text before references) to address potential concerns where
% appropriate. Topics include, but are not limited to, studies that involve human
% subjects, practices to data set releases, potentially harmful insights,
% methodologies and applications, potential conflicts of interest and sponsorship,
% discrimination/bias/fairness concerns, privacy and security issues, legal
% compliance, and research integrity issues (e.g., IRB, documentation, research
% ethics). This statement should not be more than 1 page.

\subsection*{Reproducibility statement}

We provide the information needed to reproduce PORTER and our experiments
throughout the main paper and appendix. The complete method, including
requirement support, ISE, marginal residency selection, and Push/Pull
reconciliation, is specified in Secs.~\ref{sec:method_support}--\ref{sec:method_reconcile},
with derivations and theoretical properties in
Appendix~\ref{app:ise_derivation}. Experimental protocols, metrics, and
comparison settings are described in Sec.~\ref{sec:exp_setup} and the
corresponding experiment sections. Implementation details, default
hyperparameters, upstream 3DSG configurations, and residency-policy definitions
are provided in Appendix~\ref{app:implementation}; hyperparameter sensitivity
is reported in Appendix~\ref{app:parameter_sensitivity}. Real-robot sensing,
ground-truth annotation, task sequencing, and edge--cloud transfer details are
given in Appendix~\ref{app:real_robot_details}. Computational complexity and
runtime measurement protocols are documented in
Appendices~\ref{app:porter_complexity} and~\ref{app:runtime_details}. We plan
to release the implementation and experiment configurations upon publication.



\bibliography{iclr2027_conference}
\bibliographystyle{iclr2027_conference}

\clearpage
\appendix
\section{Appendix}
\subsection{Derivation and Properties of Irreplaceable Support Erasure}
\label{app:ise_derivation}

We derive the marginal Irreplaceable Support Erasure (ISE) used in
Sec.~\ref{sec:method_ise} and summarize several properties that clarify its
behavior.

\paragraph{Requirement-wise support drop.}
Recall that the Noisy-OR coverage of requirement $r$ under residency set $S$ is
\begin{equation}
    c_r(S)
    =
    1-\prod_{u\in\mathcal V}\bigl(1-e_{ur}(S)\bigr).
\end{equation}
For $v\in S$, offloading $P_v$ changes only its effective support from
$a^F_{vr}$ to $a^A_{vr}$. Let
\[
    Q_{vr}(S)
    =
    \prod_{u\neq v}\bigl(1-e_{ur}(S)\bigr).
\]
Then
\begin{align}
    c_r(S)
    &=
    1-\bigl(1-a^F_{vr}\bigr)Q_{vr}(S),\\
    c_r(S\setminus\{v\})
    &=
    1-\bigl(1-a^A_{vr}\bigr)Q_{vr}(S),
\end{align}
and hence
\begin{align}
    \delta_{vr}(S)
    &=
    c_r(S)-c_r(S\setminus\{v\})\\
    &=
    \bigl(a^F_{vr}-a^A_{vr}\bigr)
    \prod_{u\neq v}\bigl(1-e_{ur}(S)\bigr).
\end{align}
The first factor is the support removed with $P_v$, while the second is the
portion of requirement $r$ not already covered by alternative objects. Thus,
the same payload incurs little loss when substitutes remain and becomes more
valuable as those substitutes disappear.

\paragraph{Marginal ISE.}
From Eq.~\ref{eq:porter_ise},
\begin{equation}
    \operatorname{ISE}_q(S)
    =
    \sum_{r\in\mathcal R_q}
    \omega_r
    \log_2
    \frac{c_r(\mathcal V)+\epsilon_{\mathrm{ISE}}}
         {c_r(S)+\epsilon_{\mathrm{ISE}}}.
\end{equation}
The marginal erasure caused by offloading $P_v$ is
\begin{align}
    \Delta_q^{\mathrm{ISE}}(v;S)
    &=
    \operatorname{ISE}_q(S\setminus\{v\})
    -
    \operatorname{ISE}_q(S)\\
    &=
    \sum_{r\in\mathcal R_q}
    \omega_r
    \log_2
    \frac{c_r(S)+\epsilon_{\mathrm{ISE}}}
         {c_r(S\setminus\{v\})+\epsilon_{\mathrm{ISE}}}.
\end{align}
The full-residency reference $c_r(\mathcal V)$ therefore cancels in the
marginal: it defines the set-level erasure baseline, but the next offloading
decision depends only on the current residency state.

Using
$c_r(S\setminus\{v\})=c_r(S)-\delta_{vr}(S)$ gives
\begin{equation}
    \Delta_q^{\mathrm{ISE}}(v;S)
    =
    \sum_{r\in\mathcal R_q}
    \omega_r
    \log_2
    \left(
        \frac{c_r(S)+\epsilon_{\mathrm{ISE}}}
             {c_r(S)-\delta_{vr}(S)+\epsilon_{\mathrm{ISE}}}
    \right).
\end{equation}

\paragraph{Set-function structure.}
\textbf{Proposition.}
Assume $0\leq a^A_{vr}\leq a^F_{vr}<1$ and $\omega_r\geq0$.
Then, for every requirement $r$, the Noisy-OR coverage $c_r(S)$ is a
monotone submodular function of the resident set $S$. Moreover, the normalized
retained log-support utility
\begin{equation}
    F_q(S)
    :=
    \sum_{r\in\mathcal R_q}
    \omega_r
    \log_2
    \frac{c_r(S)+\epsilon_{\mathrm{ISE}}}
         {c_r(\emptyset)+\epsilon_{\mathrm{ISE}}}
\end{equation}
is nonnegative, normalized, monotone, and submodular. Consequently,
\begin{equation}
    \operatorname{ISE}_q(S)
    =
    F_q(\mathcal V)-F_q(S)
\end{equation}
is monotone non-increasing and supermodular.

\textit{Proof.}
For a fixed requirement $r$, define
\[
    \lambda_{vr}
    :=
    \frac{1-a^F_{vr}}{1-a^A_{vr}}
    \in (0,1],
    \qquad
    \kappa_r
    :=
    \prod_{u\in\mathcal V}
    \bigl(1-a^A_{ur}\bigr).
\]
Because an object contributes $a^F_{vr}$ when resident and $a^A_{vr}$
otherwise, the uncovered support can be written as
\begin{equation}
    1-c_r(S)
    =
    \kappa_r
    \prod_{u\in S}\lambda_{ur}.
\end{equation}
Hence, for $v\notin S$,
\begin{equation}
    c_r(S\cup\{v\})-c_r(S)
    =
    \kappa_r
    (1-\lambda_{vr})
    \prod_{u\in S}\lambda_{ur}.
\end{equation}
Since every $\lambda_{ur}\leq1$, for any $A\subseteq B$ and
$v\notin B$,
\begin{equation}
    c_r(A\cup\{v\})-c_r(A)
    \geq
    c_r(B\cup\{v\})-c_r(B),
\end{equation}
which establishes diminishing returns; monotonicity follows because the
marginal gain is nonnegative. Thus, $c_r$ is monotone submodular.

Now let
\[
    \Delta c_r(v\mid S)
    :=
    c_r(S\cup\{v\})-c_r(S).
\]
The marginal gain of the corresponding log-support term is
\begin{align}
    \Delta F_r(v\mid S)
    &=
    \log_2
    \frac{c_r(S\cup\{v\})+\epsilon_{\mathrm{ISE}}}
         {c_r(S)+\epsilon_{\mathrm{ISE}}} \\
    &=
    \log_2\!\left(
        1+
        \frac{\Delta c_r(v\mid S)}
             {c_r(S)+\epsilon_{\mathrm{ISE}}}
    \right).
\end{align}
For $A\subseteq B$, submodularity of $c_r$ gives
$\Delta c_r(v\mid A)\geq\Delta c_r(v\mid B)$, while monotonicity gives
$c_r(A)\leq c_r(B)$. Therefore,
\[
    \frac{\Delta c_r(v\mid A)}
         {c_r(A)+\epsilon_{\mathrm{ISE}}}
    \geq
    \frac{\Delta c_r(v\mid B)}
         {c_r(B)+\epsilon_{\mathrm{ISE}}},
\]
and hence
$\Delta F_r(v\mid A)\geq\Delta F_r(v\mid B)$.
Each log-support term is therefore monotone submodular, and so is their
nonnegative weighted sum $F_q$.

Finally, since $\operatorname{ISE}_q(S)=F_q(\mathcal V)-F_q(S)$,
ISE is the complement of a monotone submodular retained-support utility and
is therefore monotone non-increasing and supermodular.
\hfill$\square$

Equivalently, for $A\subseteq B$ and $v\in A$,
\begin{equation}
    \Delta_q^{\mathrm{ISE}}(v;A)
    \geq
    \Delta_q^{\mathrm{ISE}}(v;B),
\end{equation}
so removing a payload can only become more costly as the resident set shrinks.
This formally captures PORTER's state-dependent repricing: substitutes reduce
marginal erasure while they remain available, and the surviving payloads become
progressively harder to remove as that redundancy disappears.

\paragraph{Relation to submodular knapsack optimization.}
The preceding proposition gives an equivalent retained-utility view of the
PORTER objective. Since
\[
    \operatorname{ISE}_q(S)
    =
    F_q(\mathcal V)-F_q(S),
\]
minimizing ISE under the local memory budget is equivalent to
\begin{equation}
    S_q^\star
    \in
    \arg\max_{S\subseteq\mathcal V}
    F_q(S)
    \qquad
    \text{s.t.}
    \qquad
    \sum_{v\in S} b_v \leq B.
\end{equation}
Because $F_q$ is monotone submodular and the payload footprint is modular,
the target-residency problem is an instance of monotone submodular
maximization under a knapsack constraint. Classical forward-selection
algorithms for this problem admit constant-factor approximation guarantees;
in particular, a $(1-1/e)$ approximation is achievable with the standard
enumeration-assisted greedy construction~\citep{sviridenko2004note}.

PORTER does not use this forward construction directly. Instead, it starts
from the currently complete resident set and repeatedly removes the payload
with the smallest marginal ISE per freed byte,
as in Eq.~\ref{eq:porter_reverse_greedy}. This reverse-greedy form directly
produces an eviction ordering and naturally supports progressive memory
release as the budget tightens. Although it exploits the same diminishing-returns structure, the classical $(1-1/e)$ guarantee for forward
submodular-knapsack maximization does not directly apply to this removal rule,
and we do not claim such a guarantee for the current scheduler.

\paragraph{Consistency with sequential offloading.}
Consider a removal sequence
$S_0=\mathcal V\supset S_1\supset\cdots\supset S_T$, where
$S_t=S_{t-1}\setminus\{v_t\}$. The marginal erasures telescope:
\begin{align}
    \sum_{t=1}^{T}
    \Delta_q^{\mathrm{ISE}}(v_t;S_{t-1})
    &=
    \operatorname{ISE}_q(S_T)
    -
    \operatorname{ISE}_q(\mathcal V)\\
    &=
    \operatorname{ISE}_q(S_T),
\end{align}
since $\operatorname{ISE}_q(\mathcal V)=0$. Thus, sequential marginal erasure
is consistent with set-level ISE.

\paragraph{Local interpretation.}
For a small requirement-wise drop $\delta_{vr}(S)$,
\begin{equation}
    \log_2
    \frac{c_r(S)+\epsilon_{\mathrm{ISE}}}
         {c_r(S)-\delta_{vr}(S)+\epsilon_{\mathrm{ISE}}}
    \approx
    \frac{\delta_{vr}(S)}
         {(c_r(S)+\epsilon_{\mathrm{ISE}})\ln 2}.
\end{equation}
Locally, ISE can therefore be interpreted as irreplaceable support loss
weighted by requirement fragility: the same absolute loss receives a larger
penalty when the available support is already weak. This distinguishes ISE
from a linear coverage-loss objective.
\subsection{Computational Complexity and Runtime Scalability}
\label{app:porter_scalability}

We complement the runtime results in Sec.~\ref{sec:exp_runtime} with the
theoretical complexity of PORTER and detailed latency measurements. The former
characterizes how residency reasoning scales with scene size, while
the latter verifies this behavior on the Tier~3 scene graphs used in our
experiments.

\subsubsection{Theoretical Complexity}
\label{app:porter_complexity}

PORTER's computation follows the two task-conditioned stages of the method:
constructing requirement support from persistent anchors and selecting the
local residency set using marginal ISE. We analyze these stages after task
decomposition and text embeddings are available. Let $N=|\mathcal V|$ be the
number of objects in the persistent 3DSG, $M=|\mathcal R_q|$ the number of
task requirements, $d$ the embedding dimension, and $K$ the number of
candidate supporters retained per requirement.

\paragraph{Requirement support.}
Because every object retains its lightweight anchor locally, PORTER can score a
new task without accessing its heavy payloads. Computing the similarity
$s_{vr}$ between all $N$ object anchors and $M$ requirement embeddings costs
\begin{equation}
    T_{\mathrm{support}}
    =
    \mathcal O\!\bigl(NM(d+\log K)\bigr),
    \label{eq:porter_support_complexity}
\end{equation}
where the $\log K$ term accounts for retaining the top-$K$ supporters of each
requirement. For fixed $K$, this is dominated by the $NMd$ similarity
computation.

More importantly, top-$K$ support bounds the number of objects that can have
nonzero marginal ISE. Let
\begin{equation}
    \mathcal C_q
    =
    \bigcup_{r\in\mathcal R_q}\mathcal K_r,
    \qquad
    C=|\mathcal C_q|.
    \label{eq:porter_active_set}
\end{equation}
Then
\begin{equation}
    C\leq \min(N,KM).
    \label{eq:porter_candidate_bound}
\end{equation}
Objects outside $\mathcal C_q$ have zero support for every requirement under
Eq.~\ref{eq:porter_support}, and therefore zero marginal ISE. Thus, although
task support requires scanning all $N$ persistent anchors, only the
support-active set $\mathcal C_q$ enters the nontrivial residency decisions.

\paragraph{Marginal ISE selection.}
For a current residency set $S$, evaluating
$\Delta_q^{\mathrm{ISE}}(v;S)$ for one candidate requires aggregating over the
$M$ requirements and therefore costs $\mathcal O(M)$. If $L\leq C$ active
payloads are removed before the memory budget is satisfied, reverse greedy
evaluates at most $C-\ell$ candidates at removal step $\ell$. Hence,
\begin{align}
    T_{\mathrm{ISE}}
    &=
    \mathcal O\!\left(
        M\sum_{\ell=0}^{L-1}(C-\ell)
    \right)\\
    &=
    \mathcal O(LCM)
    \subseteq
    \mathcal O(C^2M).
    \label{eq:porter_ise_complexity}
\end{align}

Combining requirement support and marginal ISE selection gives
\begin{equation}
    T_{\mathrm{PORTER}}
    =
    \mathcal O\!\bigl(
        NM(d+\log K)+LCM
    \bigr).
    \label{eq:porter_total_complexity}
\end{equation}
Since $C\leq KM$ and $L\leq C$, the ISE-selection term is independent of
scene size $N$. Under PORTER's fixed top-$K$ construction, with bounded task
requirements and a fixed embedding dimension, the overall complexity with
respect to scene size reduces to
\begin{equation}
    T_{\mathrm{PORTER}}=\mathcal O(N).
    \label{eq:porter_linear_complexity}
\end{equation}
Thus, scene growth increases only the cost of searching persistent anchors,
while state-dependent ISE reasoning remains bounded by the task-conditioned
support set, preserving linear scaling as persistent scene memory grows.

\subsubsection{Runtime Measurements}
\label{app:runtime_details}

\paragraph{Protocol.}
We measure PORTER on the ConceptGraphs representations used in the
Tier~3 experiments. Cubicle, Apartment, and Office contain 229, 429, and 922
objects and contribute 11, 8, and 6 tasks, respectively, yielding 25
task-switch cases. The tasks contain 55 requirement instances over
42 distinct labels, with $M\in\{2,3\}$ and CLIP embedding dimension
$d=1024$. Tier~3 oracle object labels are used directly as requirements and
their OpenCLIP text embeddings are precomputed, isolating PORTER's residency
computation from external language-model inference. Timings are collected on
one pinned logical CPU of an AMD EPYC~7542.

\begin{table}[ht]
    \centering
    \small
    \caption{
    Stage-wise PORTER latency over 25 Tier~3 task-switch cases on one pinned
    logical CPU. Values are median and 95th-percentile latency.
    }
    \label{tab:porter_runtime_breakdown}
    \setlength{\tabcolsep}{5pt}
    \begin{tabular}{lcc}
        \toprule
        Stage & Median (ms) & P95 (ms) \\
        \midrule
        Requirement support       & 0.716 & 1.412 \\
        Marginal ISE selection    & 0.963 & 1.612 \\
        Support + ISE             & 1.686 & 3.007 \\
        Push/Pull reconciliation  & 0.304 & 0.859 \\
        \midrule
        \textbf{PORTER core}      & \textbf{2.034} & \textbf{3.938} \\
        \bottomrule
    \end{tabular}
\end{table}

We separately measure requirement-support construction, marginal ISE selection,
and Push/Pull reconciliation. Subtotal and total statistics are computed from
the jointly measured per-invocation latency before aggregation; consequently,
their medians need not equal the sum of independently reported stage medians.
Table~\ref{tab:porter_runtime_breakdown} reports the resulting stage-wise and
PORTER-core latencies.

\paragraph{Scaling with scene size.}
Table~\ref{tab:porter_runtime_scene} breaks down PORTER-core latency by scene.
As the number of persistent objects increases from 229 to 429 and 922, median
latency rises from $1.679$\,ms to $2.197$\,ms and $3.492$\,ms, respectively,
while P95 remains below $4.4$\,ms in all three scenes.
These measurements provide an empirical counterpart to
Eq.~\ref{eq:porter_linear_complexity}: increasing persistent scene size
primarily increases anchor scoring, while the task-conditioned ISE reasoning
remains small.

\begin{table}[ht]
    \centering
    \small
    \caption{
    PORTER-core latency across Tier~3 scenes of different sizes.
    }
    \label{tab:porter_runtime_scene}
    \setlength{\tabcolsep}{6pt}
    \begin{tabular}{lrrr}
        \toprule
        Scene & Objects $N$ & Median (ms) & P95 (ms) \\
        \midrule
        Cubicle   & 229 & 1.679 & 2.214 \\
        Apartment & 429 & 2.197 & 2.841 \\
        Office    & 922 & 3.492 & 4.324 \\
        \bottomrule
    \end{tabular}
\end{table}

\subsection{Implementation Details}
\label{app:implementation}

PORTER is implemented as a lightweight residency layer over an upstream 3DSG.
It operates on an object-level interface consisting of a persistent identity,
a vision--language embedding, the byte size of the migratable payload, and its
current residency state. The upstream builder retains control of scene
construction and object updates, while PORTER determines only which acquired
payloads should remain locally resident.

\subsubsection{PORTER Deployment and Configuration}
\label{app:porter_configuration}

\paragraph{Anchor--payload instantiation.}
For each upstream object $v$, PORTER separates a lightweight persistent anchor
$A_v$ from its builder-specific payload $P_v$. The anchor retains the object
identity and normalized vision--language embedding required for retrieval and
residency reasoning. The payload contains the complete object state required
to restore the object, including its geometric and accumulated observation
data. We define $b_v$ as the serialized size of this native payload. Anchor
storage is excluded from the migratable-payload budget because anchors remain
local under every residency policy.

\paragraph{Anchor memory scaling.}
Anchors are lightweight but not free: because one anchor is retained for every
persistent object, their total footprint grows linearly with catalog size.
In our current implementation, the dominant fixed-size anchor field is a
1024-dimensional FP32 vision--language embedding, requiring
$1024\times4=4096$ bytes ($4$ KiB) per object. The embedding buffer alone
therefore requires approximately $3.91$ MiB, $39.06$ MiB, $390.63$ MiB, and
$3.81$ GiB for $10^3$, $10^4$, $10^5$, and $10^6$ objects, respectively.
Additional identities, indexing structures, and runtime metadata introduce
further implementation-dependent overhead.

PORTER targets the substantially heavier builder-specific payload term rather
than making persistent metadata independent of scene size. Consequently, the
edge footprint remains linear in the number of persistent anchors even when
payload residency is tightly bounded. Compressing, hierarchically indexing, or
forgetting long-lived anchors is complementary to PORTER and outside the scope
of the present residency layer.

\paragraph{Update semantics for cloud-resident payloads.}
PORTER treats Push as a residency transition rather than a distributed update
protocol. Once a payload is pushed, its cloud copy remains frozen at the last
persisted state and does not receive builder-specific updates from subsequent
observations. If the payload is later pulled, PORTER restores that state to the
edge, after which the upstream 3DSG builder resumes its native object-update
procedure. Lightweight anchors remain locally available for identity and
residency reasoning throughout this interval.

This choice preserves PORTER's construction-agnostic interface: propagating
new observations into cloud-resident payloads would require builder-specific
association, fusion, and conflict-resolution rules for heterogeneous payload
representations. Such remote synchronization is complementary to PORTER but
outside the scope of the present residency layer.
Our current experiments validate restoration across task switches but do not
isolate objects that change while their payloads are cloud-resident. Measuring
staleness under long-horizon dynamic scenes, and combining PORTER with
builder-specific delta logging or cloud-side update mechanisms, are therefore
natural extensions.

% \paragraph{Task requirements.}
% PORTER is agnostic to how functional requirements are obtained. For the
% Tier~2 task-switching experiments, we use a GPT-5.2 model to decompose each
% task instruction into a compact set of retrievable object requirements.
% Decompositions are fixed across paired experiments so that residency decisions
% are evaluated under identical task specifications. For Tier~3, the benchmark
% already annotates the objects required by each complex instruction; we use
% these object labels directly as requirements to isolate residency reasoning
% from requirement-generation errors. Ground-truth 3D boxes remain
% evaluation-only and are never exposed to PORTER or the competing policies.
% Unless otherwise stated, requirements are weighted uniformly,
% $\omega_r=1/|\mathcal R_q|$.

\paragraph{Requirement interface and parser instantiation.}
PORTER operates on a supplied set of functional requirements and is agnostic
to how this set is obtained. For the Tier~2 experiments, we instantiate this
interface using GPT-5.2 with the fixed prompt below. The parsed requirement
set is fixed for each task and shared across all paired comparisons. Although
the parser output schema includes a \texttt{weight} field, we do not use
model-generated weights; all requirements are assigned uniformly as
$\omega_r=1/|\mathcal R_q|$, following Sec.~3.2. For Tier~3, we instead use
the benchmark-provided target labels directly as requirements, thereby
isolating residency reasoning from requirement-generation errors.
Ground-truth 3D boxes remain evaluation-only and are never exposed to PORTER
or the competing policies. We do not separately evaluate robustness to parser
errors or omitted requirements; if a necessary requirement is absent from the
supplied set, PORTER may undervalue the corresponding task support.
Requirement generation is therefore treated as an external interface rather
than part of the PORTER residency objective. For reproducibility, the fixed
Tier~2 parser prompt is:

\begin{quote}
\small
\raggedright

\textbf{System:}
Decompose a robot task into the smallest set of independently retrievable
object requirements for a 3D scene graph. Return JSON only, with schema
\texttt{\{"requirements":[\{"text":str,"weight":number\}]\}}.
Use short concrete noun phrases. Include every object or destination whose
presence is necessary to perform the task. Do not emit actions, synonyms,
explanations, or duplicate phrases. Preserve multiple requirements when the
task needs multiple objects. If the task needs exactly one object, return
exactly one requirement; never invent a second requirement to increase the
list length.

\textbf{User:}
\texttt{Task: \{task\_text\}}
\end{quote}

\paragraph{Support and ISE configuration.}
Objects and requirements are encoded in the same OpenCLIP ViT-H/14
vision--language space using 1024-dimensional normalized embeddings. Support
is constructed according to Eq.~\ref{eq:porter_support}, and residency is
ranked by the marginal ISE per byte defined in
Eq.~\ref{eq:porter_reverse_greedy}. The default configuration used
throughout the experiments is summarized in
Table~\ref{tab:porter_configuration}.

\begin{table}[ht]
    \centering
    \small
    \caption{Default PORTER configuration used in the main experiments.}
    \label{tab:porter_configuration}
    \setlength{\tabcolsep}{5pt}
    \begin{tabular}{@{}ll@{}}
        \toprule
        Component & Setting \\
        \midrule
        Vision--language encoder
            & OpenCLIP ViT-H/14 \\
        Embedding dimension
            & $d=1024$ \\
        Requirement weights
            & $\omega_r=1/|\mathcal R_q|$ \\
        Similarity threshold
            & $\alpha=0.2$ \\
        Candidate supporters
            & $K=5$ per requirement \\
        Anchor support scale
            & $\eta=0$ \\
        ISE stabilizer
            & $\epsilon_{\mathrm{ISE}}=10^{-6}$ \\
        \bottomrule
    \end{tabular}
\end{table}

% Support values are clipped away from one and Noisy-OR products are accumulated
% in the log domain for numerical stability. Marginal ISE is recomputed after
% every active removal, so the value of an object changes as its substitutes
% disappear.

% \paragraph{Track-specific operation.}
% Tier~2 invokes PORTER when the task changes and constructs a new target
% residency set before reconciling it with the placement inherited from the
% previous task. We constrain the relative support erasure using
% \begin{equation}
%     1-2^{-\operatorname{ISE}_q(S)}
%     \leq
%     \tau_{\mathrm{frac}},
%     \qquad
%     \tau_{\mathrm{frac}}=0.05,
%     \label{eq:app_tier2_guard}
% \end{equation}
% and continue reverse-greedy removal while this condition remains satisfied.
% Thus, the operating point adapts to the support structure of each task rather
% than imposing the same byte budget on every task.

% Tier~3 instead evaluates the complete compression trajectory. Each task starts
% from the same frozen Keep-All representation, and PORTER repeatedly applies
% its dynamic marginal-ISE-per-byte rule to produce a complete offloading order.
% Task performance is re-evaluated after each whole-object removal.

\subsubsection{Upstream 3DSG Baselines}
\label{app:builder_configuration}

PORTER does not alter the construction procedure of the upstream 3DSG. We use
the released implementations and recommended configurations of
ConceptGraphs~\citep{gu2024conceptgraphs},
ReasoningGraph~\citep{puigjaner2026relationship},
Clio~\citep{maggio2024clio}, and
DAAAM~\citep{gorlo2026describe}, adapting only their input interfaces where
required by JITOMA-Bench. Within each paired comparison, the native and
PORTER-augmented variants share the same perceptual input, task sequence, and
builder configuration; they differ only in payload residency.

For the controlled Tier~3 experiments, we use a single frozen ConceptGraphs
reconstruction for every residency policy, eliminating scene-construction
variation from the comparison. The resulting maps contain 229, 429, and 922
objects for Cubicle, Apartment, and Office, respectively. Builder-specific
adapters retain each method's native object representation rather than
converting all methods into a common simplified graph. Complete builder
configurations will be released with the code.

\subsubsection{Residency Policy Configuration}
\label{app:residency_baselines}

\paragraph{Common protocol.}
The Tier~3 policy comparison fixes the upstream ConceptGraphs representation
and varies only the order in which payloads are offloaded. Every method
receives the same object catalog, payload sizes, task annotations, and
vision--language embeddings, and each task begins from the identical Keep-All
state. After every removal, evaluation uses only the payload-bearing objects
that remain locally resident.

\paragraph{Random and Largest-First.}
Random removes objects according to a uniformly sampled permutation; we
average results over five seeds. Largest-First removes objects in decreasing
order of payload size $b_v$, representing a memory-driven policy without task
semantics.

\paragraph{CLIP/Byte.}
The complete task instruction is embedded as $\hat{\mathbf g}_q$, and each
object is scored by
\begin{equation}
    \phi_v^{\mathrm{clip}}
    =
    \frac{
        \hat{\mathbf f}_v^{\mathsf T}\hat{\mathbf g}_q
    }{b_v}.
    \label{eq:app_clip_byte}
\end{equation}
Objects with the lowest scores are removed first. CLIP/Byte therefore combines
direct task relevance with payload cost but does not explicitly reason over
functional requirements.

\paragraph{Requirement/Byte.}
We first form a weighted task embedding from the individual requirement
embeddings,
\begin{equation}
    \bar{\mathbf g}_q
    =
    \sum_{r\in\mathcal R_q}
    \omega_r\hat{\mathbf g}_r,
    \qquad
    \phi_v^{\mathrm{req}}
    =
    \frac{
        \hat{\mathbf f}_v^{\mathsf T}\bar{\mathbf g}_q
    }{b_v}.
    \label{eq:app_requirement_byte}
\end{equation}
Objects with the lowest scores are again removed first. This baseline adds
functional task decomposition but does not model substitute coverage,
fragility, or state-dependent repricing.

\paragraph{PORTER and controlled ablations.}
PORTER removes the payload with the smallest current
$\Delta_q^{\mathrm{ISE}}(v;S)/b_v$ and recomputes the marginal after every
removal. The component ablations in
Table~\ref{tab:porter_component_ablation} each modify one element of this
policy: Noisy-OR aggregation, log-survival calibration, dynamic re-ranking, or
byte normalization. Unless explicitly varied, all variants use the default
parameters in Table~\ref{tab:porter_configuration}. The support-set
sensitivity study varies only
$K\in\{1,3,5,10,15,20\}$.

For a target byte-offloading ratio $\rho_B$, all methods are evaluated at the
first whole-object state whose cumulative removed payload reaches that target,
\begin{equation}
    t_\pi(\rho_B)
    =
    \min\left\{
        t:
        \frac{B_{\mathrm{off}}^\pi(t)}
             {B(\mathcal V)}
        \geq \rho_B
    \right\}.
    \label{eq:app_integral_budget}
\end{equation}
Payloads are never fractionally removed; consequently, the realized ratio may
slightly exceed the target by one object, and the same rule is applied to
every policy.

\subsubsection{Hardware and Software Environment}
\label{app:hardware_setup}

Experiments are conducted on a server equipped with AMD EPYC~7542 CPUs,
1\,TiB of host memory, and NVIDIA RTX~A6000 GPUs. Upstream 3DSG construction
uses GPU acceleration where required by the corresponding builder, whereas
the PORTER residency core runs on CPU.
\subsection{Hyperparameter Sensitivity}
\label{app:parameter_sensitivity}

\subsubsection{Support Set Size K}
\label{app:topk_sensitivity}

In Eq.~\ref{eq:porter_support}, $K$ controls the maximum number of candidate
supporters retained for each task requirement. Small $K$ yields a sparser
object--requirement support graph but may omit useful substitutes, whereas
larger $K$ admits more alternative support at increased computation.
We follow the same Tier~3 progressive-compression protocol as
Sec.~\ref{sec:exp_scheduler} and vary
$K\in\{1,3,5,10,15,20\}$ while keeping all other settings fixed.
We report relative mR@3 together with Overall and Tail nAUC, using the same
definitions as in Sec.~\ref{sec:exp_ablation}. The main experiments use
$K=5$.

\begin{table*}[ht]
    \centering
    \small
    \caption{
    Sensitivity of \textsc{Porter} to the number of candidate supporters $K$
    on JITOMA-Bench Tier~3. Overall nAUC covers the complete compression
 and Tail nAUC covers $[95\%,100\%]$. All remaining columns
    report mR@3 relative to Keep-All. ${}^\dagger$ denotes the default
    setting used in the experiments.
    }
    \label{tab:topk_sensitivity}
    \setlength{\tabcolsep}{6pt}
    \begin{tabular}{@{}crrrrrrr@{}}
        \toprule
        & \multicolumn{2}{c}{Curve nAUC (\%) $\uparrow$}
        & \multicolumn{5}{c}{
            Relative $\mathrm{mRecall@3}$ at $\rho_B$ (\%) $\uparrow$
        } \\
        \cmidrule(lr){2-3}
        \cmidrule(lr){4-8}
        $K$
        & Overall
        & Tail
        & $95\%$
        & $97\%$
        & $99\%$
        & $99.5\%$
        & $99.9\%$ \\
        \midrule
        1
        & 61.19 & 35.85
        & 39.34 & 39.34 & 36.07 & 26.23 & 0.00 \\

        3
        & 98.41 & 72.84
        & 85.25 & 85.25 & 54.10 & 50.82 & 31.15 \\

        $5^\dagger$
        & 98.30 & 72.03
        & 85.25 & 81.97 & 59.02 & 50.82 & 31.15 \\

        10
        & 98.26 & 73.14
        & 85.25 & 85.25 & 63.93 & 60.66 & 31.15 \\

        15
        & 98.22 & 73.27
        & 85.25 & 85.25 & 63.93 & 50.82 & 9.84 \\

        20
        & 98.23 & 73.49
        & 85.25 & 90.16 & 63.93 & 50.82 & 9.84 \\
        \bottomrule
    \end{tabular}
\end{table*}

\paragraph{Results.}
Table~\ref{tab:topk_sensitivity} reports the sensitivity of \textsc{Porter} to $K$.
With $K=1$, Overall and Tail nAUC fall to $61.19\%$ and $35.85\%$,
respectively. With only one candidate per requirement, Noisy-OR has essentially
no alternative support from which to infer replaceability, consistent with the
large degradation observed when Noisy-OR is removed in
Table~\ref{tab:porter_component_ablation}.

Once multiple alternatives are retained, PORTER is largely insensitive to the
precise value of $K$. Across $K\in\{3,5,10,15,20\}$, Overall nAUC varies by
only $0.19$ points and Tail nAUC by $1.46$ points. No single setting dominates every extreme checkpoint because payloads are removed as whole objects rather than fractional bytes, so small changes in object ordering can produce discrete jumps near full offloading.
This stability indicates that the observed gains are not driven by a
particular top-$K$ truncation once several alternative supporters are retained.
It does not, however, rule out settings with unusually diffuse long-tail
support, for which adaptive or threshold-based support sets may be preferable.
We use $K=5$ as a
moderate fixed setting that preserves multiple alternative support while keeping the
task-conditioned candidate set compact.

\subsubsection{Similarity Threshold}
\label{app:alpha_sensitivity}

In Eq.~\ref{eq:porter_support}, $\alpha$ filters weak
object--requirement matches before support aggregation. We follow the same
Tier~3 progressive-compression protocol as Sec.~\ref{sec:exp_scheduler}, fix
$K=5$, and vary $\alpha\in\{0.10,0.15,0.20,0.25,0.30\}$ while keeping all
other settings fixed. As in the main experiments, mR@3 is pooled across Tier~3 tasks at each
checkpoint and normalized by the pooled Keep-All reference. We report
Overall and Tail nAUC using the definitions in
Sec.~\ref{sec:exp_ablation}. The main experiments use $\alpha=0.20$.

\begin{table}[ht]
    \centering
    \small
    \caption{
    \textbf{Sensitivity of PORTER to the similarity threshold $\alpha$ on
    JITOMA-Bench Tier~3.}
    Overall nAUC covers the complete compression curve and Tail nAUC covers
    $[95\%,100\%]$. Remaining columns report mR@3 relative to Keep-All.
    $\dagger$ denotes the default setting.
    }
    \label{tab:alpha_sensitivity}
    \setlength{\tabcolsep}{4.0pt}
    \begin{tabular}{@{}crrrrrrr@{}}
        \toprule
        $\alpha$ & Overall & Tail & 95\% & 97\% & 99\% & 99.5\% & 99.9\% \\
        \midrule
        0.10 & 98.55 & 71.86 & 90.16 & 81.97 & 59.02 & 45.90 & 31.15 \\
        0.15 & 98.04 & 71.48 & 85.25 & 81.97 & 59.02 & 50.82 & 31.15 \\
        0.20$^\dagger$ & 98.30 & 72.03 & 85.25 & 81.97 & 59.02 & 50.82 & 31.15 \\
        0.25 & 96.58 & 60.05 & 77.05 & 59.02 & 50.82 & 50.82 & 31.15 \\
        0.30 & 62.75 & 30.93 & 34.43 & 34.43 & 29.51 & 24.59 & 0.00 \\
        \bottomrule
    \end{tabular}
\end{table}

\paragraph{Results.}
PORTER is stable over moderate similarity thresholds. For
$\alpha\in\{0.10,0.15,0.20\}$, Overall nAUC varies by only $0.51$ points and
Tail nAUC by $0.55$ points. At the default $\alpha=0.20$, each requirement
retains $4.78$ nonzero supporters on average, with no zero-support
requirements. Increasing $\alpha$ to $0.25$ reduces this average to $3.29$
and leaves $2/55$ requirements without positive support, while Tail nAUC drops
to $60.05\%$. At $\alpha=0.30$, $31/55$ requirements have zero support and
Overall and Tail nAUC fall to $62.75\%$ and $30.93\%$. These results indicate
that PORTER is insensitive within a broad moderate range, but degrades once
thresholding removes too much substitute structure. We use $\alpha=0.20$ as
a moderate default that filters weak matches while retaining sufficient
alternative support for replaceability reasoning.

\subsection{Real-Robot Experimental Details}
\label{app:real_robot_details}

\subsubsection{Platform and Sensing}

\begin{figure}[ht]
    \centering
    \includegraphics[width=\linewidth]
    {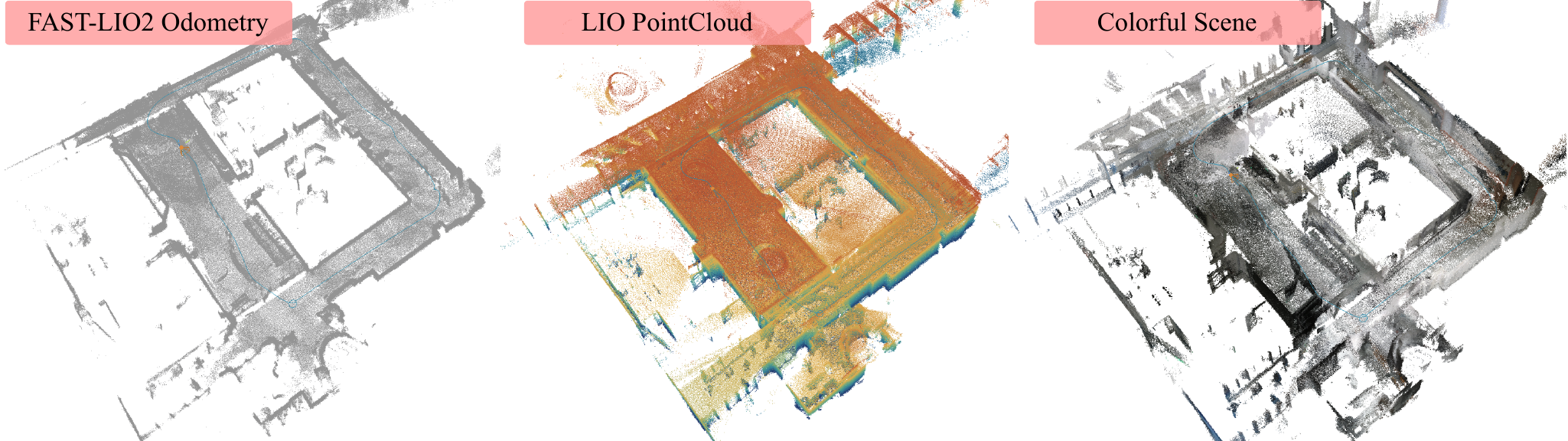}
    \caption{
    \textbf{Dual-LiDAR odometry and scene reconstruction.}
    The two Livox Mid-360 streams are fused through FAST-LIO2 \citep{xu2022fast} to estimate the
    robot trajectory and register LiDAR observations in a common frame.
    The resulting geometry provides a dense scene-level reference that is
    subsequently aligned with the RealSense observations.
    }
    \label{fig:real_robot_lio}
\end{figure}

The real-robot experiment uses an AgileX mobile platform equipped with an
Intel RealSense D435 RGB-D camera and two Livox Mid-360 LiDARs. The two LiDAR
streams are fused in our FAST-LIO2 \citep{xu2022fast} pipeline to provide a common odometric
trajectory and registered LiDAR geometry over the full traversal. This
dual-LiDAR configuration improves spatial coverage around the robot and
provides the geometric reference used throughout the experiment.
Figure~\ref{fig:real_robot_lio} visualizes the resulting odometry, registered
LiDAR point cloud, and colorized scene reconstruction.

The experiment contains 1,183 RGB-D frames collected during a continuous
traversal of the indoor environment shown in
Fig.~\ref{fig:real_robot_setup}. ConceptGraphs receives the same image sequence
and camera trajectory in both the Vanilla and PORTER conditions. PORTER
therefore changes only payload residency; the sensing trajectory and upstream
perception input are held fixed.

\subsubsection{Geometry and Ground-Truth Annotation}

\begin{figure}[ht]
    \centering
    \includegraphics[width=\linewidth]{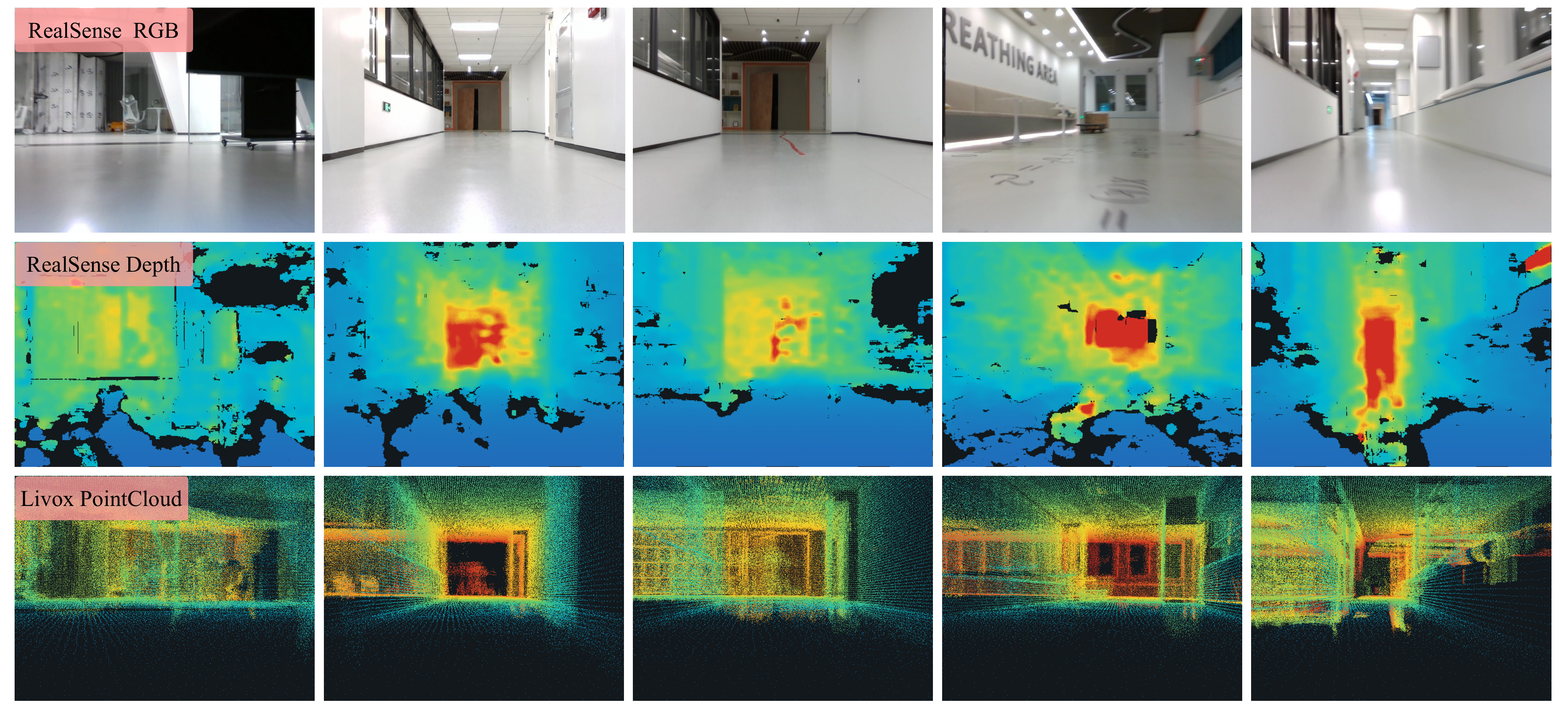}
    \caption{
    \textbf{RGB-D and LiDAR observations in the real-robot experiment.}
    Representative RealSense RGB images (top), raw D435 depth measurements
    (middle), and registered Livox point clouds (bottom). The raw depth contains
    substantial missing and noisy regions, motivating the use of LiDAR geometry
    as a higher-fidelity reference for depth refinement and 3D annotation.
    }
    \label{fig:real_robot_geometry}
\end{figure}

Although the LiDAR pipeline provides accurate global geometry and camera
registration, ConceptGraphs operates on RGB-D observations. Raw RealSense
depth is sufficiently noisy and incomplete to make precise 3D reconstruction
and object-box annotation difficult, particularly for distant surfaces and
object boundaries. We therefore use the registered LiDAR reconstruction as a
higher-fidelity geometric reference for refining the D435 depth.
Figure~\ref{fig:real_robot_geometry} compares representative RGB, raw depth,
and LiDAR observations.

Using the calibrated camera poses, the registered LiDAR reconstruction is
reprojected into each camera view to obtain LiDAR-refined depth aligned with
the RGB observations. The refined RGB-D sequence is then used to construct a
geometrically consistent ConceptGraphs representation.

\begin{figure}[ht]
    \centering
    \includegraphics[width=\linewidth]
    {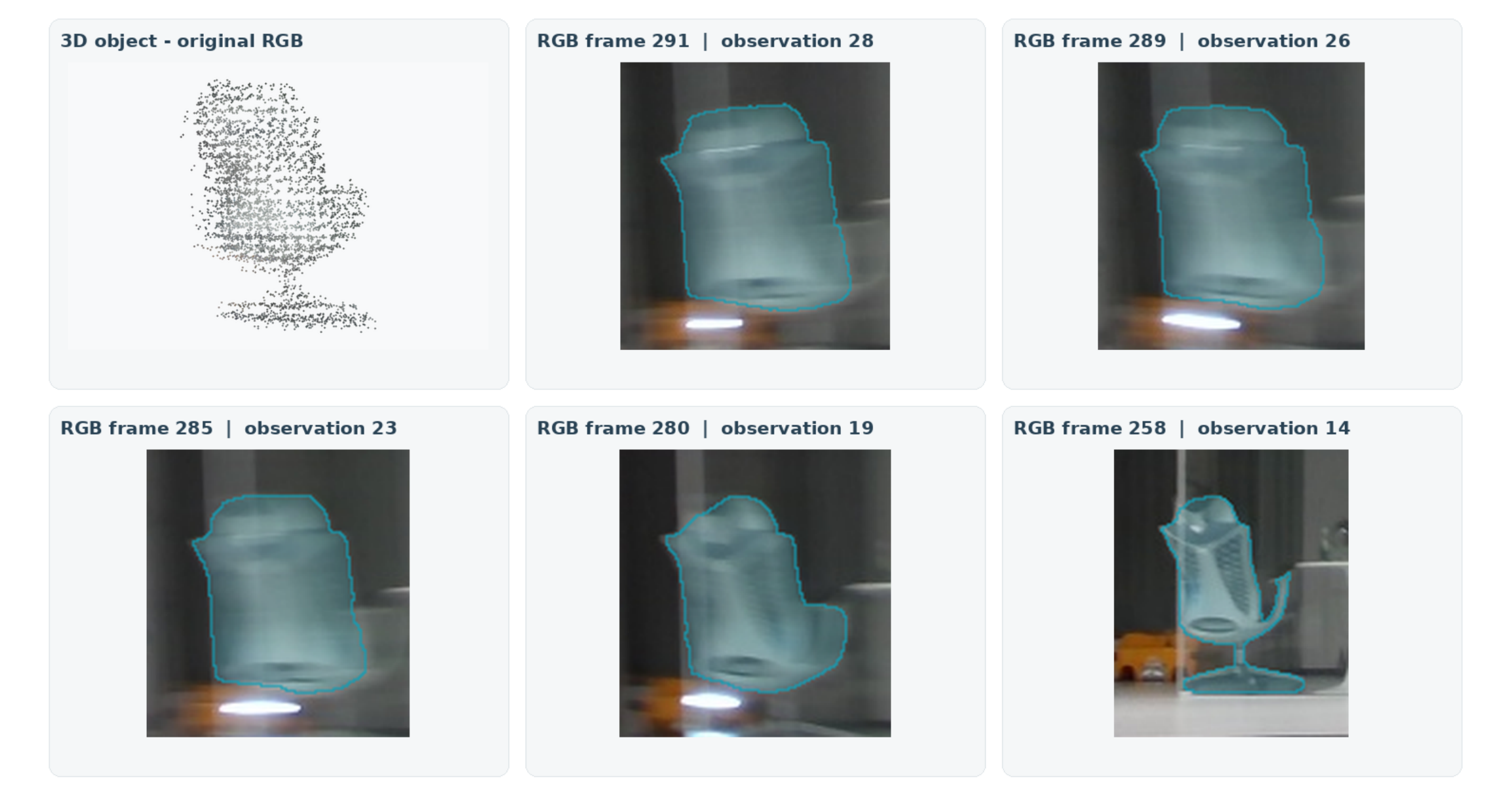}
    \caption{
\textbf{Task-object ground-truth annotation.}
Example reconstructed 3D object (left) and associated RGB observations from
multiple frames. Annotators use the accumulated views to verify object identity
and the reconstructed geometry to assign its 3D ground-truth box.
}
    \label{fig:real_robot_gt}
\end{figure}

Task-relevant ground truth is annotated at the object level. For each target,
we jointly inspect its reconstructed 3D object and the associated historical
RGB observations accumulated by ConceptGraphs, as illustrated in
Fig.~\ref{fig:real_robot_gt}. Multiple views provide additional evidence for
object identity when a single observation is ambiguous, while the reconstructed
3D geometry determines the corresponding evaluation box.

The resulting boxes serve only as evaluation ground truth and are never
provided to ConceptGraphs, PORTER, or the residency policies during operation.
Task performance is evaluated with mR@3 using the same IoU thresholds
$\{0.1,0.2,0.3\}$ as in the main benchmark experiments.

\subsubsection{Task-Switch Sequence}

We evaluate four consecutive tasks over the same continuously growing scene
memory. Their temporal boundaries and required objects are summarized in
Table~\ref{tab:real_robot_tasks}. The instructions are:

\smallskip
\noindent\textbf{T1:} Check whether the television screen is black, then find
a chair for a guest to sit on.\\
\textbf{T2:} Grasp the door handle and open the door.\\
\textbf{T3:} Pick up the table-like folding camping cot and move it away from
the wooden box.\\
\textbf{T4:} Check again whether the television screen is black, then close
the door.
\smallskip

The repeated television and door requirements test whether scene knowledge
useful to an earlier task can be recovered after its payload leaves local
residency.

\begin{table}[ht]
    \centering
    \small
    \caption{Task sequence used in the real-robot experiment.}
    \label{tab:real_robot_tasks}
    \setlength{\tabcolsep}{4pt}
    \begin{tabular}{@{}ccl@{}}
        \toprule
        Task & Frames & Required objects \\
        \midrule
        T1 & $0$--$300$
           & television, chair \\
        T2 & $305$--$600$
           & door / door handle \\
        T3 & $610$--$900$
           & camping cot, wooden box \\
        T4 & $910$--$1180$
           & television, door \\
        \bottomrule
    \end{tabular}
\end{table}

\subsubsection{Edge--Cloud Residency and Transfer}
\label{app:real_robot_transfer}

The robot and remote server maintain separate local and cloud payload stores.
The remote server establishes an SSH reverse tunnel to the robot, through which
payloads are transferred using HTTP PUT and GET requests. Before transfer,
payloads are gzip-compressed and contain the builder-specific object state,
including the object point cloud and historical observation crops.

Transfers are executed concurrently with mapping. Consequently, the mapping
TPF reported in Sec.~\ref{sec:real_robot} measures the online mapping path
rather than serializing communication into the per-frame latency. PORTER
accounts for payload volume through its marginal-ISE-per-byte criterion.
Table~\ref{tab:real_robot_system} summarizes the observed Push/Pull activity
and the resulting system-level gains.

\begin{table}[ht]
    \centering
    \small
    \caption{
    \textbf{Real-robot residency and system performance.}
    (a) Push/Pull activity and gzip-compressed payload volume across task
    switches. (b) Comparison of vanilla ConceptGraphs (CG) and
    ConceptGraphs+PORTER.
    }
    \label{tab:real_robot_system}
    \vspace{2pt}

    \begin{minipage}[t]{0.58\linewidth}
        \centering
        \textbf{(a) Edge--cloud transfer}
        \vspace{2pt}

        \footnotesize
        \setlength{\tabcolsep}{2.4pt}
        \begin{tabular}{@{}crrrr@{}}
            \toprule
            Switch
            & Push
            & Pull
            & Upload
            & Download \\
            & & & (MB) & (MB) \\
            \midrule
            T1 & 18  & 0 & 1.42   & 0.00 \\
            T2 & 131 & 0 & 166.08 & 0.00 \\
            T3 & 203 & 4 & 189.14 & 3.72 \\
            T4 & 186 & 9 & 182.22 & 19.23 \\
            \midrule
            Total & 538 & 13 & 538.86 & 22.95 \\
            \bottomrule
        \end{tabular}
    \end{minipage}
    \hfill
    \begin{minipage}[t]{0.39\linewidth}
        \centering
        \textbf{(b) System summary}
        \vspace{2pt}

        \footnotesize
        \setlength{\tabcolsep}{3pt}
        \begin{tabular}{@{}lrr@{}}
            \toprule
            Metric & CG & +PORTER \\
            \midrule
            Frames
                & 1183 & 1183 \\
            Avg. local objs.
                & 278.58 & 114.07 \\
            Peak local objs.
                & 540 & 224 \\
            Final local objs.
                & 531 & 224 \\
            TPF (s/frame)
                & 6.276 & 1.863 \\
            mR@3
                & 75.0\% & 75.0\% \\
            \bottomrule
        \end{tabular}
    \end{minipage}
\end{table}

\paragraph{Transfer latency and connectivity.}
Push and Pull have asymmetric roles in the online system. Push transfers move
payloads that are no longer required by the current task and therefore execute
asynchronously off the mapping-critical path. Pull transfers are task-critical,
since newly required cloud-resident payloads must be restored before they can
be used. In our real-robot deployment, Push traffic is substantially larger:
the four task switches upload $538.86$\,MB in total, whereas only
$22.95$\,MB is downloaded across 13 Pulls.

Table~\ref{tab:real_robot_transfer_latency} reports the measured end-to-end
transfer times under the deployment network. The large Pushes at T2--T4
transfer $166.08$--$189.14$\,MB in $9.12$--$11.02$\,s while mapping
continues concurrently. In contrast, the task-critical Pulls are much smaller:
T3 restores $3.72$\,MB in $0.76$\,s and T4 restores $19.23$\,MB in
$1.24$\,s. Thus, the large communication volume is dominated by asynchronous
eviction, while the synchronous restoration path transfers only the subset
required by the new task.

\begin{table}[ht]
    \centering
    \small
    \caption{
    \textbf{Measured edge--cloud transfer latency in the real-robot deployment.}
    Times are end-to-end measurements under the deployment network.
    }
    \label{tab:real_robot_transfer_latency}
    \setlength{\tabcolsep}{5pt}
    \begin{tabular}{@{}crrrr@{}}
        \toprule
        Switch &
        Upload (MB) &
        Push (s) &
        Download (MB) &
        Pull (s) \\
        \midrule
        T1 & 1.42   & 0.54  & 0.00  & --   \\
        T2 & 166.08 & 9.12  & 0.00  & --   \\
        T3 & 189.14 & 11.02 & 3.72  & 0.76 \\
        T4 & 182.22 & 10.06 & 19.23 & 1.24 \\
        \bottomrule
    \end{tabular}
\end{table}

These measurements characterize our deployment network rather than a
network-independent latency guarantee. Physical transfer time depends on
payload size, available bandwidth, RTT, and contention. Under intermittent
connectivity, locally resident payloads and all lightweight anchors remain
available, but a cloud-resident payload cannot be restored until connectivity
recovers. PORTER therefore preserves logical scene memory across a network
outage, while immediate access to offloaded payloads remains subject to the
availability of the backing store.

The nonzero Pull operations in T3 and T4 are particularly important: they show
that cloud-resident payloads remain reusable across task switches and can be
restored when their task value rises again. The experiment therefore exercises
the full residency cycle---local retention, Push, persistent cloud storage,
and subsequent Pull---rather than treating offloading as irreversible deletion.

\paragraph{Switching cost.}
PORTER deliberately optimizes task-conditioned information retention rather
than jointly optimizing residency and network migration. The target residency
set is therefore independent of the current residency; the latter determines
only the Push/Pull needed to realize that target. This separation avoids
coupling ISE to deployment-specific bandwidth, RTT, and asymmetric transfer
costs. Although our experiments invoke PORTER at task switches, the method does
not require a residency update at every transition: its invocation policy is
external to ISE and can be adapted to task duration, memory pressure, network
conditions, or application-specific batching. Rapidly alternating tasks may
therefore use a coarser residency schedule, while migration-aware hysteresis or
switching penalties can be added at the execution layer without changing the
ISE criterion.

\subsubsection{Real-Robot Summary}

As shown in Table~\ref{tab:real_robot_system}, PORTER reduces average and peak
local residency by $59.1\%$ and $58.5\%$, respectively, while preserving the
same $75.0\%$ mR@3. The smaller working set also reduces mapping TPF from
$6.276$\,s/frame to $1.863$\,s/frame, a $70.3\%$ reduction, demonstrating a
system-level benefit beyond memory footprint alone.
\subsection{Comparison with Redundancy-Aware Residency Baselines}
\label{app:strong_baselines}

The main progressive-compression experiment focuses on a diagnostic progression
from task-agnostic removal to relevance- and requirement-aware policies. We
further compare PORTER with two stronger set-selection baselines that explicitly
model redundancy: maximal marginal relevance (MMR) and task-weighted facility
location. All methods use the same frozen ConceptGraphs representation, oracle
task requirements, object embeddings, and payload sizes, and are evaluated
under the same Tier~3 progressive-compression protocol. At each requested
payload-byte offloading ratio, we evaluate the first whole-object state whose
cumulative removed payload reaches that ratio, without fractional-object
interpolation.

A requirement-aware submodular coverage objective is already evaluated as the
\emph{w/o Log-Information Loss} ablation in
Table~\ref{tab:porter_component_ablation}. Under the same dynamic removal rule,
this ablation is equivalent to Submodular Coverage/Byte, so we do not duplicate
it here. Instead, the following baselines provide complementary tests of
semantic diversity and set-level representativeness.

\paragraph{MMR.}
We adapt maximal marginal relevance to object residency using the same object
and requirement embeddings as the other Tier~3 policies. Let
$\mathbf v_i$ and $\mathbf q_r$ denote individually $\ell_2$-normalized object
and requirement vision--language embeddings. We define dense task relevance
and retained-set redundancy as
\begin{equation}
    R(i)
    =
    \frac{1}{M}
    \sum_{r=1}^{M}
    \mathbf v_i^\top \mathbf q_r,
    \qquad
    D(i,S)
    =
    \begin{cases}
        0, & S=\emptyset,\\
        \max_{j\in S}\mathbf v_i^\top\mathbf v_j,
        & S\neq\emptyset.
    \end{cases}
\end{equation}
Starting from an empty retained set, MMR repeatedly selects
\begin{equation}
    i^\star
    =
    \arg\max_{i\notin S}
    \left[
        \lambda R(i)
        -
        (1-\lambda)D(i,S)
    \right],
\end{equation}
with a fixed $\lambda=0.5$ for all tasks. Both terms use raw signed cosine
similarities, without clipping, support thresholding, Top-$K$ pruning, or other
PORTER-specific calibration. Ties are resolved by canonical object ID. We
construct the complete forward retention ordering and reverse it to obtain the
offloading sequence. MMR scores are therefore updated during forward
selection, as in the standard formulation, rather than recomputed as deletion
marginals during offloading. We do not byte-normalize the MMR objective;
payload sizes enter only through the common byte-offloading checkpoints used
for evaluation.

\paragraph{Task-weighted Facility-Location/Byte.}
We also evaluate a redundancy-aware representativeness objective that does not
use PORTER's Noisy-OR aggregation. Each object $u$ receives the task-conditioned
weight
\begin{equation}
    w_u
    =
    \sum_{r\in\mathcal R_q}
    \omega_r a^F_{ur},
\end{equation}
and object--object similarity is defined in the shared vision--language space as
\begin{equation}
    k(u,v)
    =
    \max\!\left(0,\hat{\mathbf f}_u^\top\hat{\mathbf f}_v\right).
\end{equation}
The retained-set utility is
\begin{equation}
    F_{\mathrm{FL}}(S)
    =
    \sum_{u\in\mathcal V}
    w_u
    \max_{v\in S} k(u,v),
\end{equation}
where the maximum is zero for $S=\emptyset$. At each removal step, the policy
selects
\begin{equation}
    v^\star
    =
    \arg\min_{v\in S}
    \frac{
        F_{\mathrm{FL}}(S)
        -
        F_{\mathrm{FL}}(S\setminus\{v\})
    }{b_v},
\end{equation}
and recomputes the marginal after every removal. Facility Location therefore
maintains a compact set representative of task-relevant objects while
discounting semantically redundant representatives, but it does not explicitly
preserve coverage separately for each functional requirement.

\begin{table}[t]
    \centering
    \small
    \caption{
    \textbf{Comparison with redundancy-aware residency baselines on
    JITOMA-Bench Tier~3.}
    Entries report pooled relative mR@3 at matched cumulative payload-byte
offloading ratios. Requirement-aware
    Submodular Coverage/Byte is identical to the w/o Log-Information Loss
    ablation in Table~\ref{tab:porter_component_ablation} and is not duplicated.
    }
    \label{tab:strong_residency_baselines}
    \setlength{\tabcolsep}{4.0pt}
    \begin{tabular}{@{}lrrrrrrr@{}}
        \toprule
        Method
        & 85\%
        & 90\%
        & 95\%
        & 97\%
        & 99\%
        & 99.5\%
        & 99.9\% \\
        \midrule
        MMR
        & 77.05
        & 77.05
        & 62.30
        & 52.46
        & 34.43
        & 19.67
        & 0.00 \\
        Task-weighted Facility-Location/Byte
        & 95.08
        & 90.16
        & 72.13
        & 72.13
        & \textbf{59.02}
        & 36.07
        & 0.00 \\
        \textbf{PORTER}
        & \textbf{100.00}
        & \textbf{100.00}
        & \textbf{85.25}
        & \textbf{81.97}
        & \textbf{59.02}
        & \textbf{50.82}
        & \textbf{31.15} \\
        \bottomrule
    \end{tabular}
\end{table}

\paragraph{Results.}
Table~\ref{tab:strong_residency_baselines} shows that PORTER's advantage is not
limited to comparisons with independent relevance heuristics. MMR explicitly
trades task relevance against semantic redundancy, while Facility Location
maintains a representative task-relevant subset. Both remain substantially
more robust than simple relevance-based removal as compression increases, but
the separation from PORTER grows once the resident set becomes scarce. At
$85\%$ offloading, PORTER retains $100\%$ of Keep-All mR@3, compared with
$95.08\%$ for Facility Location and $77.05\%$ for MMR; at $95\%$, the
corresponding values are $85.25\%$, $72.13\%$, and $62.30\%$.

Together with the w/o Log-Information Loss ablation in
Table~\ref{tab:porter_component_ablation}, these comparisons separate several
forms of redundancy reasoning. MMR promotes generic semantic diversity,
Facility Location preserves set-level representativeness, and linear
requirement coverage explicitly models functional substitutes. PORTER further
weights coverage loss by the fragility of the surviving requirement support.
This difference is most visible in the compression tail: at $99.5\%$
offloading, PORTER retains $50.82\%$ of Keep-All mR@3, compared with
$40.98\%$ for linear requirement coverage, $36.07\%$ for Facility Location,
and $19.67\%$ for MMR. At $99.9\%$, PORTER retains $31.15\%$, while MMR and
Facility Location reach zero. These results are consistent with redundancy
modeling explaining much of the robustness at moderate compression, while
fragility-aware requirement protection becomes increasingly useful after the
easily removable redundancy has been exhausted.
\subsection{Case Analysis: Residency And Task-Aligned Abstraction In CLIO}
\label{app:case_study}
% PORTER's push/pull residency strategy affords the system two complementary advantages. First, it allows the the edge-resident 3DSG payload footprint to be bounded against a hard budget, substantially improving the system's adaptability to long-term, everyday operation. A side benefit comes almost for free: by pushing large amounts of task-irrelevant information to the cloud, the local working set naturally adapts to the granularity demanded by the tasks, greatly reducing the noise faced by downstream retrieval. PORTER's memory benefit is builder-agnostic, whereas its retrieval benefit is only visible in task-driven builders—making Clio \cite{maggio2024clio}, whose Information Bottleneck clustering explicitly performs task-directed retrieval, a natural showcase for both.

PORTER primarily serves as a residency manager, reducing the edge-resident working set while preserving downstream task capability as shown in Figure \ref{fig:porter_policy_curve}. Beyond this primary memory benefit, reshaping the local working set can also reduce the interference faced by downstream object retrieval. We use Clio~\citep{maggio2024clio}, a task-driven 3DSG builder, to make this interaction particularly explicit: changes in local residency are directly reflected in the set and composition of object candidates exposed to the task. We illustrate two representative outcomes below—one in which PORTER recovers an initially failed object retrieval, and another in which an already-correct retrieval is preserved under substantial compression.

\subsubsection{Case 1: find spice bottles}
\begin{figure}[ht]
    \centering
    \includegraphics[width=\linewidth]{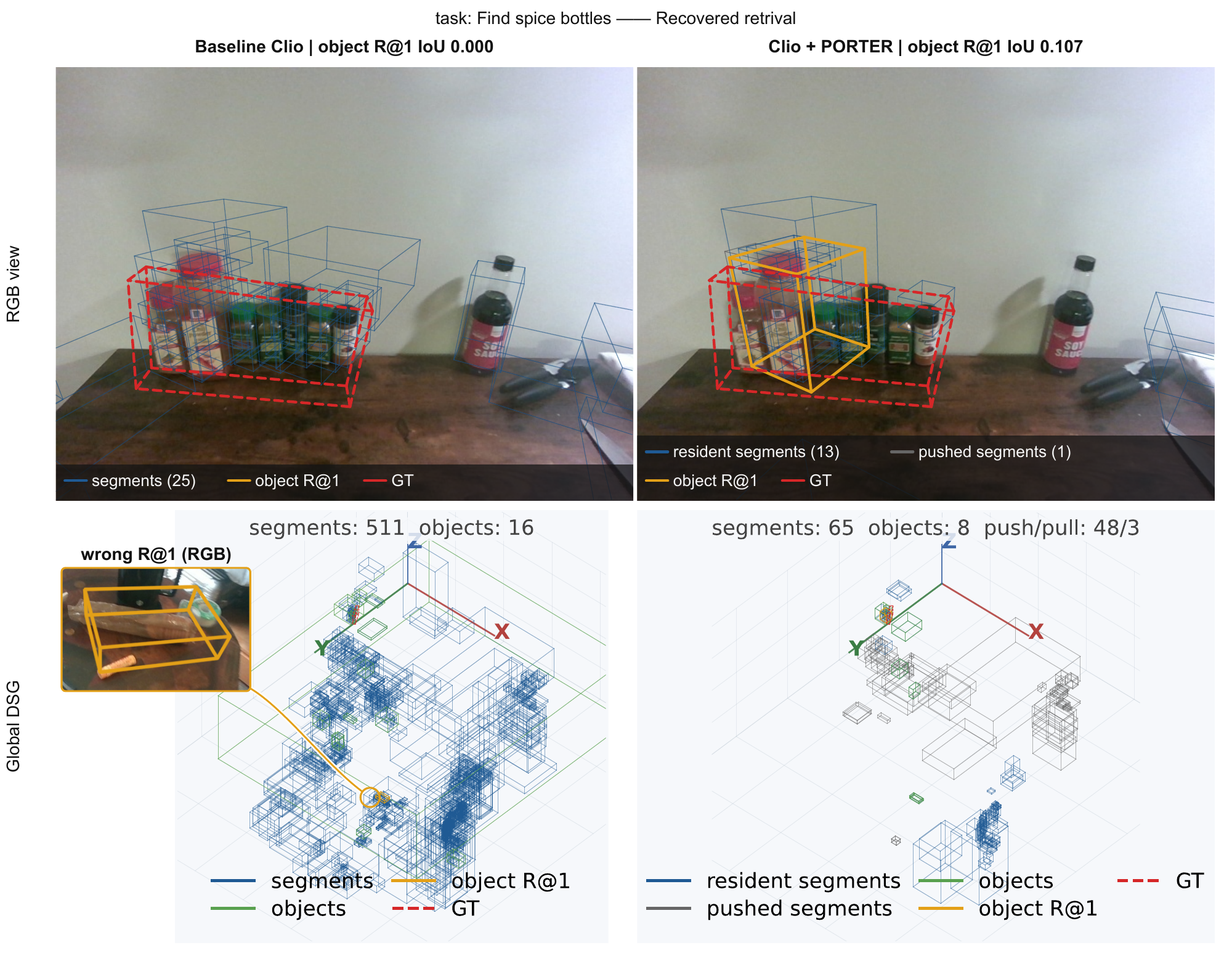}
    \caption{
    \textbf{Visualization of the find spice bottles case.} Clio first organizes the scene segments (blue box) into object abstractions (green box) using its task-driven clustering, and retrieves the object with the highest CLIP similarity to the task query, highlighted in yellow; the red dashed box denotes the ground truth. PORTER does not directly manage objects, but instead controls the residency of their underlying segments by push/pull operations: resident segments and pulled segments remain blue, while pushed segments between task injection and query are shown in gray, and an object becomes locally unavailable when its constituent segments are offloaded. The top row shows a local RGB view of these structures, whereas the bottom row shows the corresponding global 3DSG context. The inset on the lower left gives the RGB appearance and global location of the incorrect Top-1 object in Clio baseline. Segments/objects counts report the complete query-time 3DSG state, and push/pull records the segment-payload transfers performed between task injection and query.
    }
    \label{fig:find_spice_bottles}
\end{figure}

Figure \ref{fig:find_spice_bottles} shows the ``find spice bottles'' task in the Apartment scene. In Clio, 511 local segments are clustered into 16 objects which are then ranked by their CLIP similarity to the task query. The highest-scoring object is a compact, semantically plausible distractor, but lies outside the ground truth, yielding an IoU of 0. Meanwhile, the actual target is buried in a noisy 14-segment cluster and ranks only seventh with a CLIP similarity of \(0.2156\).
PORTER changes both sides of this competition. It reduces the local working set to 65 segments and 8 objects, pushing out part of the previously higher-ranked distracting evidence. At the same time, removing noisy segments allows the target to be re-clustered into a much tighter 2-segment object. Its CLIP similarity increases to \(0.2377\), promoting it from rank 7 to the highest-similarity Top-1 retrieval, with an IoU of \(0.107\). This case therefore illustrates PORTER's retrieval-side denoising effect: a smaller local working set suppresses distracting evidence while yielding a cleaner representation of the true target.

\subsubsection{Case 2: clean backpacks}
\begin{figure}[ht]
    \centering
    \includegraphics[width=\linewidth]{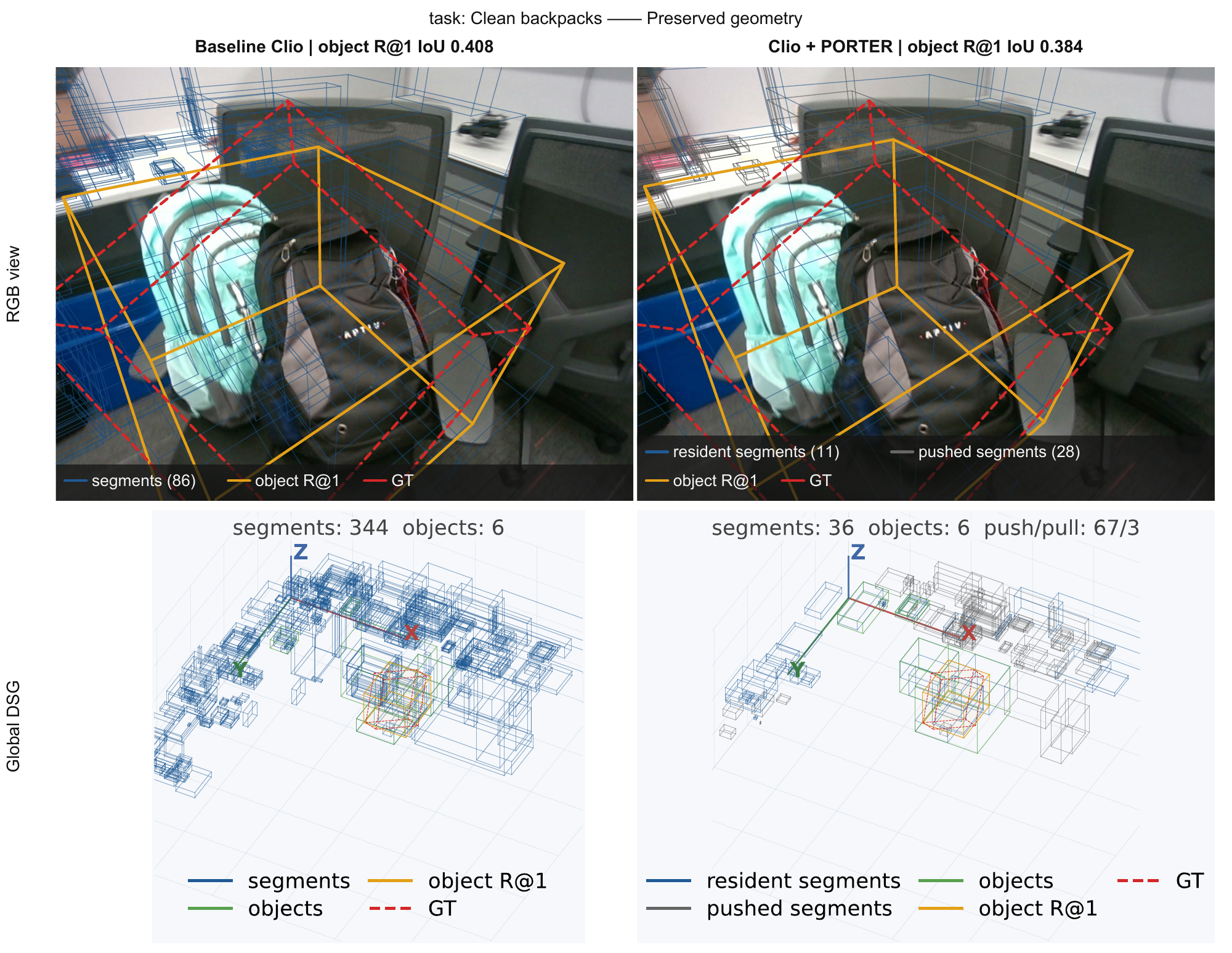}
    \caption{
    \textbf{Visualization of the clean backpacks case.} Clio clusters the blue scene segments into green object abstractions with its task-driven Information Bottleneck (IB) objective, and the yellow box marks the object with the highest CLIP similarity to the query; the red dashed box shows the ground truth. PORTER operates only on segment residency by push/pull operations: locally retained segments and pulled segments remain blue, pushed segments between task injection and query are shown in gray. The RGB panels provide a local view around the queried target, while the lower panels show the corresponding global 3DSG and the overall reduction in resident scene content. Segment/object counts are measured over the complete query-time 3DSG, and push/pull denotes the segment-payload transfers between task injection and query.
    }
    \label{fig:clean_backpacks}
\end{figure}

Figure \ref{fig:clean_backpacks} shows the complementary ``clean backpacks'' case. Clio already retrieves the target correctly, with an IoU of \(0.408\). PORTER then reduces the local working set from 344 to only 36 segments, while preserving all six object-level abstractions. The retrieved object remains aligned with the same backpack region, with an IoU of \(0.384\). In this case, PORTER removes a large amount of redundant local evidence while preserving the object representation that is already useful for retrieval.

\end{document}